\documentclass[10pt,twocolumn,letterpaper]{article}

\usepackage{cvpr}
\usepackage{times}
\usepackage{epsfig}
\usepackage{graphicx}
\usepackage{amsmath}
\usepackage{amssymb}
\usepackage{gensymb}
\usepackage{textcomp}
\usepackage{multirow}
\usepackage{mathtools,xparse}
\usepackage{enumitem}
\usepackage{booktabs}
\usepackage{microtype}
\usepackage{array}
\usepackage{makecell}
\usepackage{floatrow}

\usepackage[pagebackref=true,breaklinks=true,colorlinks,bookmarks=false]{hyperref}

\newcommand\comment[1]{}

\cvprfinalcopy 

\begin{document}

\title{All-Weather Deep Outdoor Lighting Estimation}

\author{Jinsong Zhang\textsuperscript{1*},\:Kalyan Sunkavalli\textsuperscript{$\dagger$},\:Yannick Hold-Geoffroy\textsuperscript{$\dagger$},\:Sunil Hadap\textsuperscript{$\dagger$}, \\Jonathan Eisenmann\textsuperscript{$\dagger$}, \:Jean-Fran\c{c}ois Lalonde\textsuperscript{*}\\
Universit\'e Laval\textsuperscript{*}, Adobe Research\textsuperscript{$\dagger$}\\
{\tt\footnotesize jinsong.zhang.1@ulaval.ca, \{sunkaval,holdgeof,hadap,eisenman\}@adobe.com, jflalonde@gel.ulaval.ca}\\
\footnotesize\url{http://lvsn.github.io/allweather}
\vspace{-1em}
}
\maketitle

\begin{abstract}

    We present a neural network that predicts HDR outdoor illumination from a single LDR image. At the heart of our work is a method to accurately learn HDR lighting from LDR panoramas under any weather condition. We achieve this by training another CNN (on a combination of synthetic and real images) to take as input an LDR panorama, and regress the parameters of the Lalonde-Matthews outdoor illumination model~\cite{lalonde-3dv-14}. This model is trained such that it a) reconstructs the appearance of the sky, and b) renders the appearance of objects lit by this illumination. We use this network to label a large-scale dataset of LDR panoramas with lighting parameters and use them to train our single image outdoor lighting estimation network. We demonstrate, via extensive experiments, that both our panorama and single image networks outperform the state of the art, and unlike prior work, are able to handle weather conditions ranging from fully sunny to overcast skies. 

\end{abstract}
\makeatletter
\def\blfootnote{\gdef\@thefnmark{}\@footnotetext}
\makeatother

\blfootnote{\scriptsize \textsuperscript{1} Research partly done when Jinsong Zhang was an intern at Adobe Research.}


\vspace{-1em}
\section{Introduction}
\label{sec:introduction}

Estimating outdoor illumination is critical for a number of tasks such as outdoor scene understanding, image editing, and augmented reality.  However, images are formed by illumination interacting with other scene properties like geometry and surface reflectance, thus making the inversion of this process to recover lighting a highly ill-posed problem.

Prior work on this problem has used heuristics to map image features to lighting~\cite{lalonde-ijcv-12}. Instead of using hand-crafted features, a recent approach proposed learning the mapping from image appearance to outdoor illumination~\cite{holdgeoffroy-cvpr-17} using a deep neural network. In particular, they propose a non-linear optimization scheme to fit the parameters of the Ho\v{s}ek-Wilkie (HW) HDR sky model~\cite{hosek-siggraph-12,hosek-cga-13} to SUN360---a large dataset of outdoor, low dynamic range (LDR) panoramas~\cite{xiao-cvpr-12}. From this set of panoramas with (now labeled) parameters, they extract limited field of view crops, and train a CNN to regress the HW parameters from a single crop. 

\begin{figure}[t]
    \centering
    \footnotesize
    \setlength{\tabcolsep}{1pt}
    \begin{tabular}{ccc}
    \includegraphics[width=.325\linewidth]{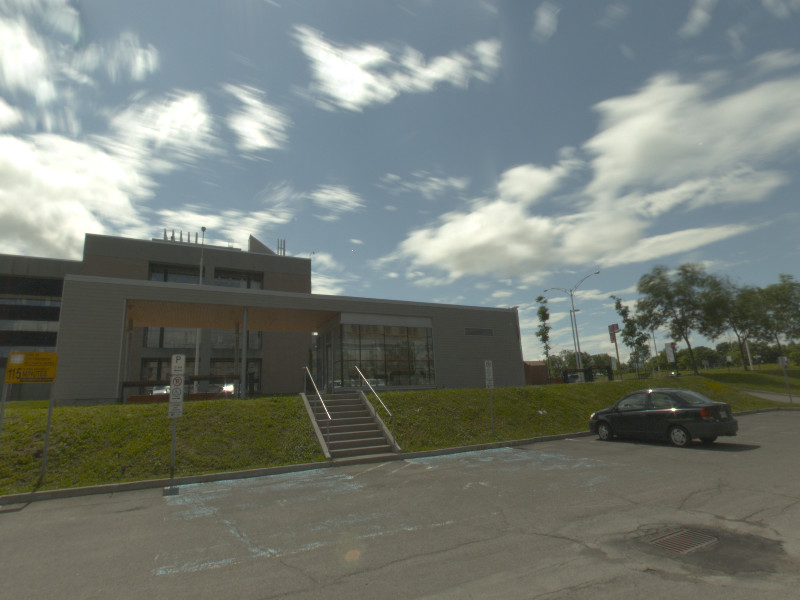} &
    \includegraphics[width=.325\linewidth]{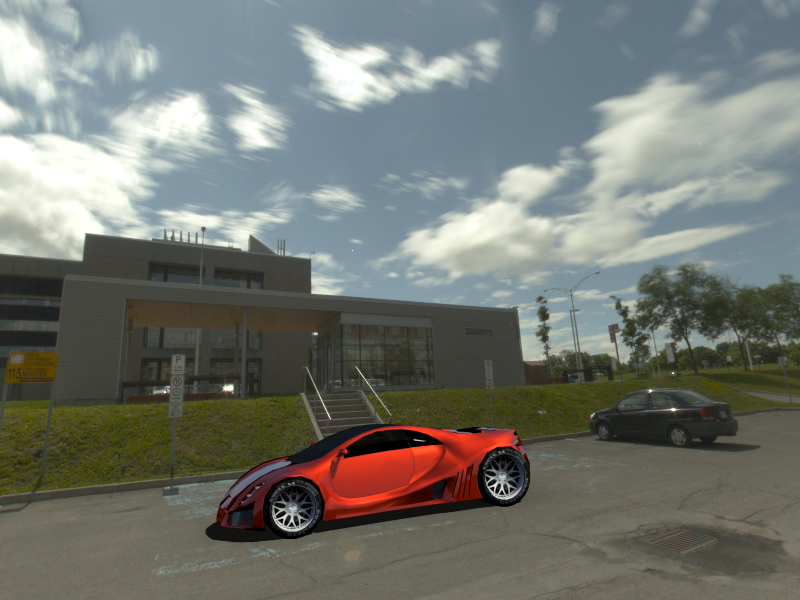}  &
    \includegraphics[width=.325\linewidth]{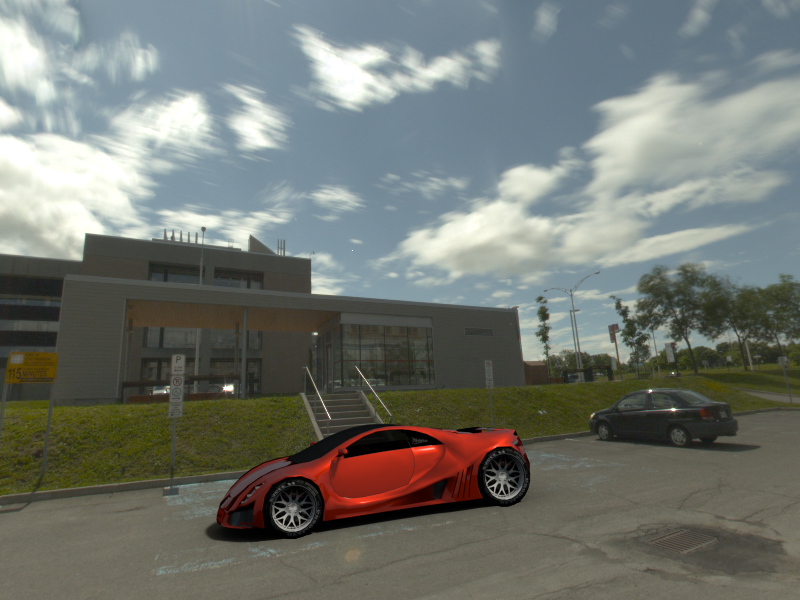} \\
    \includegraphics[width=.325\linewidth]{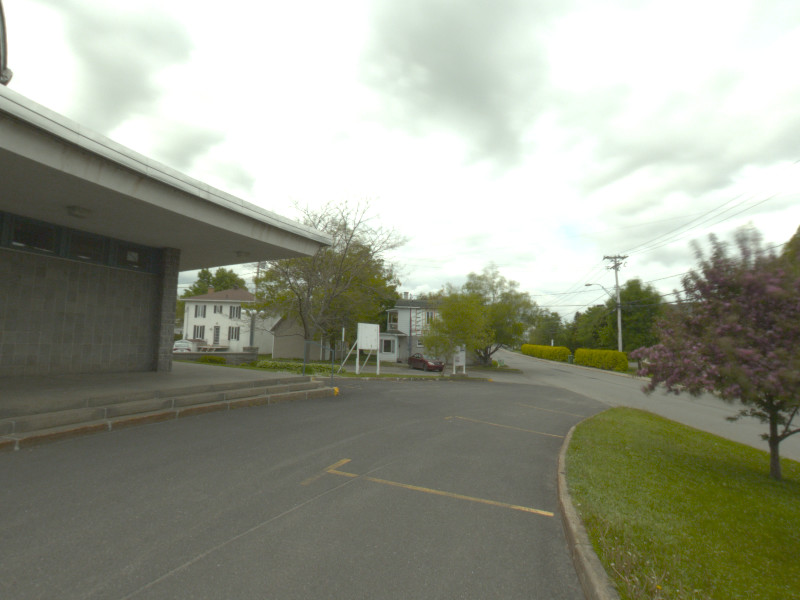} &
    \includegraphics[width=.325\linewidth]{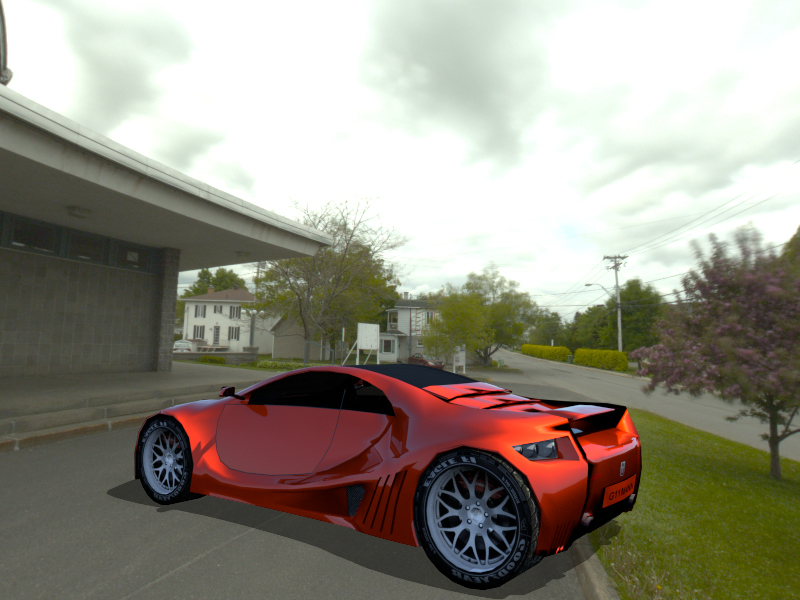}  &
    \includegraphics[width=.325\linewidth]{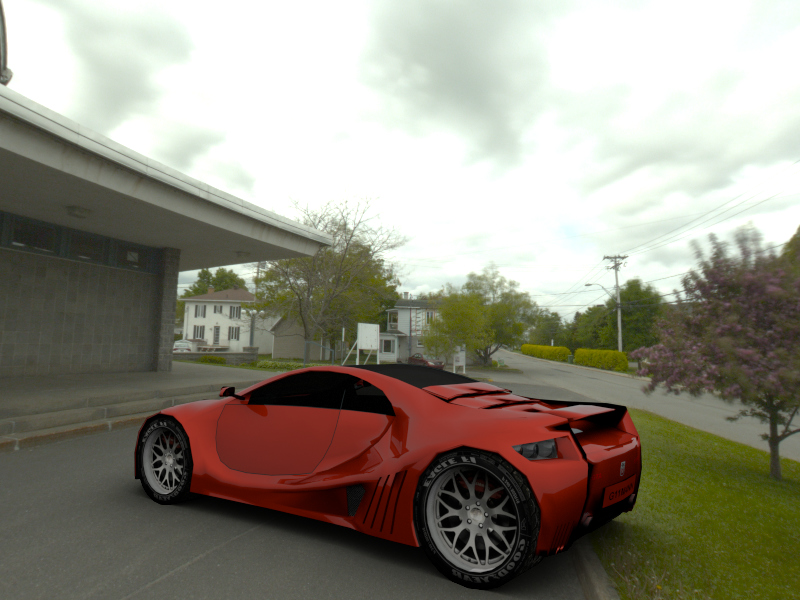} \\
    (a) Input & (b) \cite{holdgeoffroy-cvpr-17} & (c) Our lighting
    \end{tabular}
    \caption{We estimate HDR outdoor lighting from a single image (a) and use it to render a car into the scene. The state-of-the-art method of \cite{holdgeoffroy-cvpr-17} (b) produces sunny estimates for both clear (top) and overcast scenes (bottom). Our method (c) produces accurate estimates for both conditions leading to more realistic composites.
    \vspace{-2em}
    }
    \label{fig:teaser}
\end{figure}

The accuracy of \cite{holdgeoffroy-cvpr-17} thus rests on a) the ability of the HW model to represent outdoor illumination, and b) the ability to reliably fit the HW model to sky pixels in an outdoor panorama. Unfortunately, both of these steps have limitations. Indeed, the HW sky model was designed to accurately represent a \emph{subset} of possible weather conditions, specifically, completely clear skies with varying amounts of turbidity (amount of atmospheric aerosol)~\cite{kider-tog-14}. Moreover, the optimization scheme to fit the HW model to LDR panoramas can be sensitive to issues like arbitrary camera processing and outliers like patches of clouds. This leads to a clear bias towards sunny skies in their results (see fig.~\ref{fig:teaser}(b)).

Other recent approaches to estimating outdoor illumination have eschewed parametric models in favor of completely data-driven models. Zhang et al.~\cite{zhang-iccv-2017} learn to hallucinate HDR outdoor environment maps from LDR panoramas by training an encoder-decoder on synthetic data. Calian et al.~\cite{calian-eg-18} learn a data-driven sky model by training an auto-encoder on HDR sky panoramas~\cite{hdrdb,lalonde-3dv-14}, and use this representation in an inverse rendering framework to recover outdoor lighting from a single image of a face. While these learned models can better approximate outdoor lighting conditions, they conflate many of the ``intuitive'' sky parameters such as the sun position, sky color, etc. into an opaque representation that can only be decoded by a network.

In this work, we propose a method to robustly estimate a wide variety of outdoor lighting conditions from a single image. Unlike ~\cite{holdgeoffroy-cvpr-17}, we use the Lalonde-Matthews (LM) sky model~\cite{lalonde-3dv-14} that can represent a much wider set of lighting conditions, ranging from completely overcast to fully sunny. {While more expressive, the sky and sun illumination components in this model are uncorrelated; as a result, it is not possible to recover HDR sun illumination by fitting an illumination model to LDR sky pixels, as is done by \cite{holdgeoffroy-cvpr-17}. Thus, our main contribution is} a novel method to \emph{learn to label LDR panoramas with HDR lighting parameters}. Specifically, we train a network---\emph{PanoNet}---to take as input an LDR panorama, and regress the parameters of the LM model. We train PanoNet with a combination of of synthetic and real data. Moreover, because no sky model will exactly reproduce real sky pixel intensities, we show that merely learning to match \emph{sky appearance} is not sufficient for this task. Instead, we propose a novel render loss that matches the appearance of a \emph{rendered scene under the ground truth and predicted lighting}. PanoNet produces HDR lighting estimates that are significantly better than previous work. We use PanoNet to label the SUN360 dataset with lighting parameters, and similar to ~\cite{holdgeoffroy-cvpr-17}, train \emph{CropNet}---a network that regresses the lighting labels from a single crop image. Through extensive experiments (and as can be seen in fig.~\ref{fig:teaser}), we demonstrate that both PanoNet and CropNet significantly outperform the state-of-the-art, both qualitatively and quantitatively. 

\section{Related work}

{A wide array of lighting estimation methods have been presented in the literature. In this section, we focus on outdoor lighting modeling and estimation that is most related to this work.}

\vspace{-1em}
\paragraph{Outdoor lighting modeling} Modeling the solar and sky dome illumination has been extensively studied in atmospheric science, physics, and computer graphics. The Perez All-Weather model~\cite{perez1993allweather} was first introduced as an improvement over the previous CIE Standard Clear Sky model, and modeled weather variations using 5 parameters. Preetham et al.~\cite{preetham-siggraph-99} simplified this to a model with a single weather parameter---atmospheric turbidity. Ho\v{s}ek and Wilkie introduced an improvement over the Preetham model, which resulted in both a sky dome~\cite{hosek-siggraph-12} and solar disc~\cite{hosek-cga-13} analytical models. See \cite{kider-tog-14} for a comparison of how these sky models approximate clear skies on a rich outdoor dataset provided by the authors. In addition, Lalonde and Matthews~\cite{lalonde-3dv-14} proposed an empirical model for HDR skies, which they show better approximate captured skies (in RGB) under a wide variety of illumination conditions. They subsequently employ their sky model for estimating outdoor lighting from outdoor image collections. In this paper, we exploit that model (which we dub the ``LM'' model) and show that its parameters can be predicted from a single image by a CNN.

\vspace{-1em}
\paragraph{Outdoor lighting estimation} Lighting estimation from a single, generic outdoor scene was first proposed by Lalonde et al.~\cite{lalonde-ijcv-12}; they use a probabilistic combination of multiple image cues---such as cast shadows, shading, and sky appearance variation---to predict lighting. Karsch et al.~\cite{karsch2014automatic} match the input image to a large dataset of panoramas~\cite{xiao-cvpr-12}, and transfer the panorama lighting (obtained through a light classifier) to the image. However, the matching metric, that relies on image features, may not yield results that have consistent lighting. Other approaches rely on known geometry~\cite{lombardi2016reflectance} and/or strong priors on geometry and surface reflectance~\cite{barron2015shape}.

\vspace{-1em}
\paragraph{Deep learning for lighting estimation} Recent approaches have applied deep learning methods for lighting estimation. Rematas et al.~\cite{rematas2016deep} learn to infer a reflectance map (i.e., the convolution of incident illumination with surface reflectance) from a single image of an object. Subsequently, Georgoulis et al.~\cite{georgoulis2018reflectance} factor reflectance maps into lighting and material properties~\cite{Georgoulis_2017_ICCV}. Closely related to our work, Hold-Geoffroy et al.~\cite{holdgeoffroy-cvpr-17} model outdoor lighting with the parametric Ho\v{s}ek-Wilkie sky model~\cite{hosek-siggraph-12,hosek-cga-13}, and learn to estimate its parameters from a single image. As mentioned in the introduction, we take inspiration from this work and significantly improve upon it proposing a novel learning-based approach to robustly annotate LDR panoramas with different weather conditions with the parameters of the LM illumination model. This is closely related to Zhang et al.~\cite{zhang-iccv-2017} who learn to map LDR panoramas to HDR environment maps via an encoder-decoder network. Similarly, Calian et al.~\cite{calian-eg-18} {(as well as the concurrent work of Hold-Geoffroy et al.~\cite{yannick-cvpr-19})} employ a deep autoencoder to learn a data-driven illumination model. They use this learned model to estimate lighting from a face image via a multi-step non-linear optimization approach over the space of face albedo and sky parameters, that is time-consuming and prone to local minima. In contrast to the high-dimensional environment map and the learned auto-encoder representations, we use a compact and intuitive sky model---the aforementioned LM model. This allows us to easily annotate a large-scale LDR panorama dataset~\cite{xiao-cvpr-12} with lighting parameters and subsequently infer lighting from a single image of a generic outdoor scene in an end-to-end framework. Cheng et al.~\cite{cheng2018learning} estimate lighting from the front and back camera of a mobile phone. However, they represent lighting using low-frequency spherical harmonics (SH), which, as demonstrated in \cite{calian-eg-18}, does not appropriately model outdoor lighting.

\section{Brief review of the LM sky model}
\label{sec:lm-model}

In this paper, we make use of the Lalonde-Matthews (LM) sky model~\cite{lalonde-3dv-14}. This is a parameteric sun and sky model which, when fit to HDR panoramas, was determined to better approximate outdoor lighting than other, physically-based models. We now briefly summarize its form and parameters for completeness, but refer the reader to \cite{lalonde-3dv-14} for more details. 

The LM hemispherical illumination model can concisely be written as the sum of sun and sky components: 
\begin{equation}
f_\text{LM}(\mathbf{l}; \mathbf{q}_\text{LM}) = f_\text{sun}(\mathbf{l}; \mathbf{q}_\text{sun}, \mathbf{l}_\text{sun}) + f_\text{sky}(\mathbf{l};\mathbf{q}_\text{sky}, \mathbf{l}_\text{sun})\,,
\label{eqn:lm-model}
\end{equation}
where $\mathbf{l}_\text{sun} = \left[\theta_\text{sun}, \varphi_\text{sun} \right]$ is the sun position in spherical coordinates, and the $\mathbf{q}_*$ are component-specific parameters. 

The sky component $f_\text{sky}(\mathbf{l})$ in eq.~(\ref{eqn:lm-model}) is simply the Preetham sky model~\cite{preetham-siggraph-99} $f_\text{P}(\cdot)$, multiplied channel-wise with an average sky color $\mathbf{w}_\text{sky} \in \mathbb{R}^3$:
\begin{equation}
f_\text{sky}(\mathbf{l}; \mathbf{q}_\text{sky}, \mathbf{l}_\text{sun}) = \mathbf{w}_\text{sky} f_\text{P}(\theta_\text{sun}, \gamma_\text{sun}, t)\,, 
\label{eqn:sky-model}
\end{equation}
where $\gamma_\text{sun}$ is the angle between sky element $\mathbf{l}$ and the sun position $\mathbf{l}_\text{sun}$, and $t$ is the sky turbidity. The sun component $f_\text{sun}(\mathbf{l})$ in eq.~(\ref{eqn:lm-model}) is defined as
\begin{equation}
f_\text{sun}(\mathbf{l}; \mathbf{q}_\text{sun}, \mathbf{l}_\text{sun}) = \mathbf{w}_\text{sun} \exp\left(-\beta \exp\left(-\kappa / \gamma_\text{sun}\right) \right) \,, 
\label{eqn:sun-model}
\end{equation}
where $(\beta,\kappa)$ are two parameters controlling the shape of the sun, and $\mathbf{w}_\text{sun} \in \mathbb{R}^3$ is the mean sun color. In short, the LM sky model has the following 11 parameters: 
\begin{equation}
\mathbf{q}_\text{LM} = 
\left\{
\begin{array}{cccccc}
\mathbf{w}_\text{sky}, & t, & \mathbf{w}_\text{sun}, & \beta, & \kappa, & \mathbf{l}_\text{sun}
\end{array} 
\right\} \,.
\end{equation}


\begin{figure*}[!ht]
    \centering
    \includegraphics[width=\linewidth, trim={0 .9cm 4.8cm 0},clip]{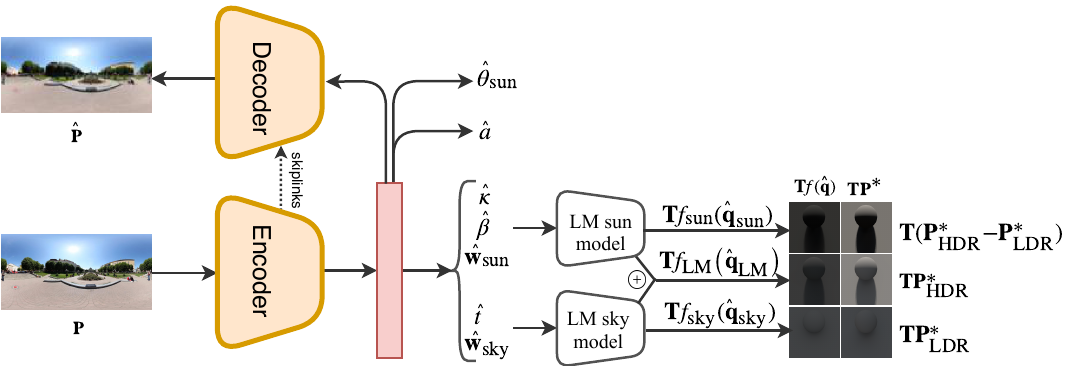}
    \caption{Architecture of PanoNet. We use an autoencoder with skip-links to regress HDR panorama from LDR. The high dynamic range lighting information is likely encoded in the bottleneck layer~\cite{zhang-iccv-2017}. From this layer we estimate the LM sky parameters by 2 FC layers. Then we use the estimated parameters to drive the LM sun and sky model and generate HDR environment maps separately. The generated sun and sky environment maps are used to render an object with a pre-computed transport matrix, $\mathbf{T}$. We compute the loss on the renders w.r.t the supervised data. Additional losses on the sun elevation $\hat\theta_\text{sun}$, hallucinated HDR sky $\mathbf{\hat{P}}$ and its render are used to force the network to encode as much lighting information in the latent vector (red). We also ask the latent vector to be insensitive to the data domain by adding the unsupervised domain adaptation branch.
	\vspace{-.5em}}
    \label{fig:panonet}
\end{figure*}
\vspace{-1em}
\section{Estimating {HDR} parametric lighting from LDR panoramas}
\label{sec:panonet}

One of the main advantages of using the Ho\v{s}ek-Wilkie model is that only its \emph{sky} parameters (sun position, turbidity and exposure)~\cite{hosek-siggraph-12} can be fit to an LDR panorama and, {because its sun and sky components are linked via the turbidity parameter}, an HDR \emph{sun} model can be extrapolated~\cite{hosek-cga-13}. This is a practical way of estimating the high dynamic range of the sun given a saturated LDR panorama~\cite{holdgeoffroy-cvpr-17}. {In this paper, we use the LM model which, because it uses \emph{independent} parameters for the sun and the sky (c.f. sec.~\ref{sec:lm-model}), is more expressive than the HW model. However, this also means its HDR sun parameters cannot be fit directly to LDR panoramas.} In this section, we therefore train a CNN that learns to predict the LM sky parameters from a single LDR panorama. 

\subsection{Architecture of PanoNet}

To regress the parameters of the LM sky model from a single LDR panorama, we take inspiration from the work of \cite{zhang-iccv-2017}, who use an autoencoder with skip-links (similar to the well-known U-net architecture~\cite{ronneberger2015u}) to regress an HDR panorama from LDR. For simplicity, they used the equirectangular format, and assumed the panorama to be rotated such that the sun is in the center. In addition, they also had another path from the latent vector $\mathbf{z}$ to two fully-connected layers that estimate the sun elevation, which was found to add robustness. Finally, they had a third path from $\mathbf{z}$ to an unsupervised domain adaptation branch that helped the network generalize to real data. 

Starting from this base architecture, we add another path from $\mathbf{z}$, this time to predict the parameters of the LM sky model. More precisely, we ask the network to learn the sun and sky intensities ($\mathbf{w}_\text{sun}$ and $\mathbf{w}_\text{sky}$ respectively), the sun shape parameters $\beta$ and $\kappa$, and the sky turbidity $t$. The new path has a structure of two consecutive FC layers with size of 512 and 256 neurons, the output layer has 9 neurons corresponding to the LM sky parameters. The resulting CNN is illustrated in fig.~\ref{fig:panonet}. 

\subsection{Loss functions}
\label{sec:panonet-loss-functions}

Several loss functions are used to train the PanoNet CNN. First, we use the same loss functions as in \cite{zhang-iccv-2017}, {namely}: 
\begin{equation}
\begin{split}
\mathcal{L}_\text{pano} &= || \mathbf{P}^* - \mathbf{\hat{P}} ||_1 \\
\mathcal{L}_\theta &= ||\theta_\text{sun}^* - \hat{\theta}_\text{sun}||_2 \\
\mathcal{L}_\text{render} &= ||\mathbf{T}\mathbf{P}^* - \mathbf{T}\mathbf{\hat{P}}||_2 \\
\mathcal{L}_\text{da} &= -\sum_{i=1} a^*_i \log \hat{a}_i \\
\end{split} \,,
\label{eqn:ldr2hdr}
\end{equation}

where symbols ($^*$) and ($\hat{\;}$) denote the ground truth and the network output respectively. From top to bottom, the various outputs of the CNN are compared to their ground truth counterparts: on the HDR panorama $\mathbf{P}$ ($\mathcal{L}_\text{pano}$), the sun elevation $\theta_\text{sun}$ ($\mathcal{L}_\theta$), a rendered image of a synthetic scene---a diffuse sphere on a plane ($\mathcal{L}_\text{render}$). {The domain adaptation branch is trained with cross-entropy loss ($\mathcal{L}_\text{da}$)}. We make the in-network rendering of the synthetic scene fast by multiplying the reconstructed panorama $\mathbf{P}$ with a pre-computed transport matrix $\mathbf{T}$ of the synthetic scene~\cite{nimeroff95rerendering}.

In addition, we also add loss functions on the estimated LM sun and sky parameters. Unfortunately, we do not have explicit targets for those parameters, so we rely on render losses exclusively: 
\begin{equation} 
\begin{split}
\mathcal{L}_\text{sky} &= ||\mathbf{T} \mathbf{P}_\text{LDR}^* - \mathbf{T}f_\text{sky}(\mathbf{\hat{q}}_\text{sky})||_2 \\
\mathcal{L}_\text{sun} &= ||\mathbf{T} (\mathbf{P}_\text{HDR}^* - \mathbf{P}_\text{LDR}^*) - \mathbf{T}f_\text{sun}(\mathbf{\hat{q}}_\text{sun})||_2 \\
\mathcal{L}_\text{LM} &= ||\mathbf{T} \mathbf{P}_\text{HDR}^* - \mathbf{T}f_\text{LM}(\mathbf{\hat{q}}_\text{LM})||_2 \\
\end{split} \,.
\label{eqn:lm}
\end{equation}
Here, we employ the same transport matrix $\mathbf{T}$ to efficiently render an image at $64\!\times\!64$ resolution, and compute L2 loss on the image. The sky-only loss $\mathcal{L}_\text{sky}$ relies on a ``ground truth'' LDR panorama $\mathbf{P}_\text{LDR}^*$, which is obtained by clipping the HDR panorama $\mathbf{P}_\text{HDR}^*$ at 1, and quantize the result to 8 bits. 
Among the render losses in eq.~\ref{eqn:lm}, we use a smaller weight for the $\mathcal{L}_\text{sky}$, which is set to 0.2.

\subsection{Datasets}

To train the PanoNet CNN, we rely on data from 5 complementary datasets. First, we employ 44,646 panoramas from the dataset of synthetic HDR panoramas from~\cite{zhang-iccv-2017}. This dataset was created by lighting a virtual 3D city model, obtained from the Unity Store, with 9,732 HDR sky panoramas from the Laval HDR sky database~\cite{hdrdb}. Second, we use 149 daytime outdoor panoramas from the HDRI Haven database~\cite{hdrihaven}. Third, we use 102 panoramas from a database of HDR outdoor panoramas~\cite{yannick-cvpr-19}.
Finally, we also train on 19,571 panoramas from the SUN360 dataset~\cite{xiao-cvpr-12}, and 4,965 from images we downloaded from Google Street View. Since these last two sets of panoramas are LDR, we only use them for the domain adaptation loss $\mathcal{L}_\text{da}$. Conversely, that loss is not evaluated when the synthetic panoramas are given as input to the network. 

\subsection{Training details}
\label{sec:panonet.details}

For training our PanoNet, we use the ADAM optimizer with a minibatch size of 80 and an initial learning rate of 0.001. Each minibatch contains 36 (45\%) synthetic HDR panoramas~\cite{zhang-iccv-2017}, 4 (5\%) captured panoramas~\cite{yannick-cvpr-19}, 4 (5\%) HDRI Haven panoramas~\cite{hdrihaven}, 4 (5\%) Google Street View images, and 32 (40\%) of SUN360 LDR panoramas~\cite{xiao-cvpr-12}. Training 500 epochs takes approximately 50 hours on an Nvidia TITAN X GPU. At test time, inference takes approximately 8ms. 

\subsection{Labeling the SUN360 panorama dataset with the PanoNet network}

Once the PanoNet CNN has been trained, we run it on all the outdoor panoramas in the SUN360 database, to obtain their estimated sun and sky LM parameters $\mathbf{\hat{q}}_\text{sun}$ and $\mathbf{\hat{q}}_\text{sky}$, respectively. {As in~\cite{holdgeoffroy-cvpr-17}, the sun position $\mathbf{l}_\text{sun}$ is obtained by finding the center of mass of the largest saturated region in the sky.} We will employ those \emph{estimates}, denoted by $(\hat{\;})$, as \emph{targets} for the second network, CropNet, whose goal will be to predict these same labels from a single limited field of view image instead of the panorama. CropNet is the subject of the next section. 

\section{Learning to estimate {HDR} parametric lighting from a single {LDR} image}
\label{sec:cropnet}

\begin{figure*}[!h]
\centering
\includegraphics[width=\linewidth, trim={0 .9cm 3.1cm 0},clip]{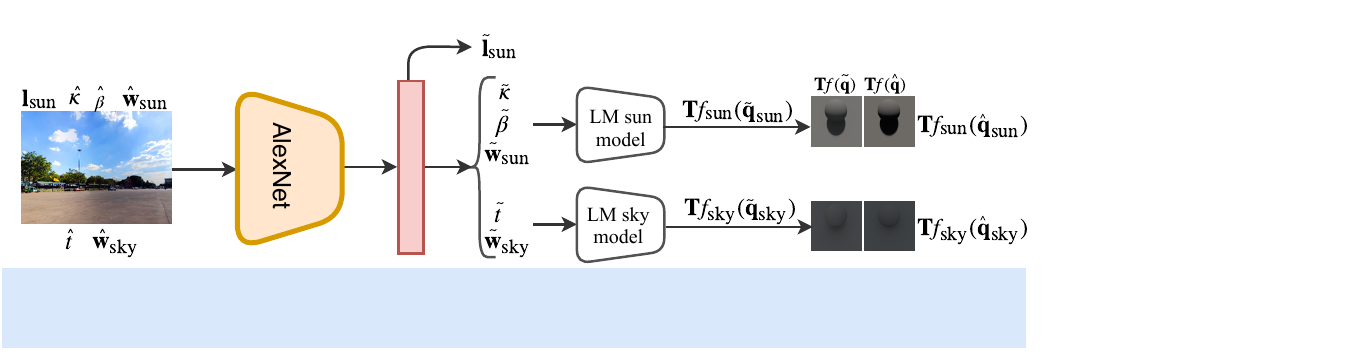}
\caption{{Architecture of CropNet}. CropNet estimates LM sky parameters from a single LDR image. All the input images are cropped from the SUN360 panoramas~\cite{xiao-cvpr-12}, which have been labeled by PanoNet (sec.~\ref{sec:panonet}) with LM sky parameters $\hat{\kappa}$, $\hat{\beta}$, $\mathbf{\hat{w}}_\text{sun}$, $\hat{t}$, $\mathbf{\hat{w}}_\text{sky}$. {The sun position $\mathbf{l}_\text{sun}$ is detected from the panorama.} We enforce a loss on these parameters, as well as on renders with the estimated lighting.}
\label{fig:cropnet.architecture}
\end{figure*}

\subsection{Architecture of CropNet}

Fig.~\ref{fig:cropnet.architecture} describes the architecture used for the CropNet CNN. Its task is to take a single LDR image as input, and estimate the LM {sun parameters ${\mathbf{l}_\text{sun}}$, $\mathbf{q}_\text{sun} = \{\mathbf{w}_\text{sun}, \beta, \kappa\}$ and the sky parameters} $\mathbf{q}_\text{sky} = \{\mathbf{w}_\text{sky}, t\}$ respectively, from that image (see sec.~\ref{sec:lm-model}). It employs a relatively straightforward architecture \`{a} la AlexNet~\cite{alexnet}, composed of 5 convolution layers, followed by two consecutive FC layers. Each convolution layer is followed by a sub-sampling step (stride of 2), batch normalization, and the ELU activation function~\cite{clevert-iclr-16} are used on all convolution layers. {The sun position branch outputs a probability distribution over the discretized sun position. As in~\cite{holdgeoffroy-cvpr-17}, we use 64 bins for azimuth and 16 bins for elevation.}  

\subsection{Loss functions}

As with PanoNet, a variety of loss functions are used to train the CropNet CNN. First, {we use a KL-divergence loss for the sun position $\mathbf{l}_\text{sun}$. Numerical loss functions are used to compare the estimated other sun/sky parameters $(\mathbf{w}_\text{sun}, \beta, \kappa, \mathbf{w}_\text{sky}, t)$} with the parameters provided by PanoNet (see sec.~\ref{sec:panonet}), $(\mathbf{\hat{q}}_\text{sun}, \mathbf{\hat{q}}_\text{sky})$. Note here that we do not have actual ``ground truth'' for those parameters---we aim for CropNet to match the predictions of PanoNet, but from a limited field of view image instead of the entire panorama. 

We begin by computing the L2 loss on all sun and sky parameters individually: 
\begin{equation}
\begin{split}
\hspace{-1em}
\begin{array}{ccc}
\mathcal{L}_\beta  = ||\hat{\beta} - \tilde{\beta}||_2 & 
\mathcal{L}_\kappa = ||\hat{\kappa} - \tilde{\kappa}||_2 & 
\mathcal{L}_t = ||\hat{t} - \tilde{t}||_2 \\
\end{array} \\
\begin{array}{cc}
\mathcal{L}_\text{wn} = ||\mathbf{\hat{w}}_\text{sun} - \mathbf{\tilde{w}}_\text{sun}||_2  &
\mathcal{L}_\text{wy} = ||\mathbf{\hat{w}}_\text{sky} - \mathbf{\tilde{w}}_\text{sky}||_2 
\end{array}
\end{split} \,,
\end{equation}
where $(\tilde{\;})$ denotes CropNet outputs. Prior to computing the loss, each parameter is normalized in the $[0,1]$ interval according to the minimum and maximum values in the training set, and the weights are all set to 1 except for $\mathcal{L}_\kappa$, $\mathcal{L}_\beta$ and $\mathcal{L}_\text{wn}$ which are set to 5, 10, and 10 respectively to balance the loss functions. 

As with PanoNet, we also employ render losses to help with the training. {Since CropNet does not have ground truth HDR lighting, two render losses are used in contrast to PanoNet. Since the sun position $\mathbf{l}_\text{sun}$ is treated independently from the other lighting parameters, we exclude the sun position from the render loss.} The following render losses are employed: 
\begin{equation} 
\begin{split}
\mathcal{L}_\text{sky} &= ||\mathbf{T}f_\text{sky}(\mathbf{\hat{q}}_\text{sky}) - \mathbf{T}f_\text{sky}(\mathbf{\tilde{q}}_\text{sky})||_2 \\
\mathcal{L}_\text{sun} &= ||\mathbf{T}f_\text{sun}(\mathbf{\hat{q}}_\text{sun}) - \mathbf{T}f_\text{sun}(\mathbf{\tilde{q}}_\text{sun})||_2 \\
\end{split} \,.
\end{equation}
In practice, this corresponds to rendering a simple scene (diffuse sphere on a plane as in sec.~\ref{sec:panonet-loss-functions}) with the estimated sky(sun)-only parameters, and comparing it with a render of that same scene with the sky(sun)-only parameters predicted from the PanoNet (c.f. sec.~\ref{sec:panonet}). The weight for both of these losses is set to 1.

\subsection{Training details}
To train our CropNet, we use the ADAM optimizer with a minibatch size of 256 and an initial learning rate of 0.001. Each minibatch contains 230 (90\%) SUN360 LDR panoramas~\cite{xiao-cvpr-12} and 26 (10\%) captured panoramas~\cite{yannick-cvpr-19}. Training 500 epochs takes approximately 90 hours on an Nvidia TITAN X GPU. At test time, inference takes approximately 25ms. 

\section{Experimental validation}

We evaluate both of our proposed CNNs on a dataset of HDR outdoor panoramas~\cite{yannick-cvpr-19} and the SUN360 LDR dataset~\cite{xiao-cvpr-12}. 
First, we show that reliable sky parameters can be estimated from LDR panoramas using our proposed PanoNet through both quantitative and qualitative comparisons with ground truth data. Then, we show how the CropNet network can robustly estimate the same sky parameters from a single LDR image.

\begin{figure*}[!t]
\scriptsize
\centering 
\setlength{\tabcolsep}{1.5pt}
\begin{tabular}{cccccc|ccccc}
\rotatebox{90}{10th} & 
\includegraphics[height=.065\linewidth]{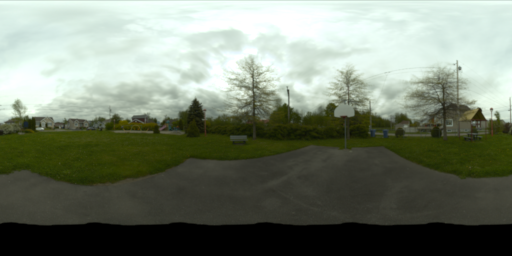}    & 
\includegraphics[height=.065\linewidth]{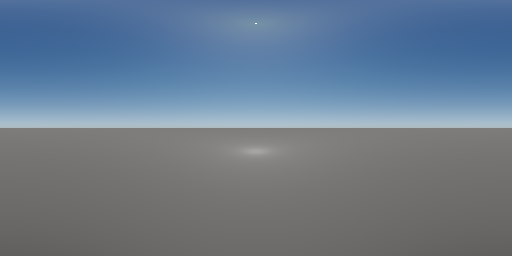}  & 
\includegraphics[height=.065\linewidth]{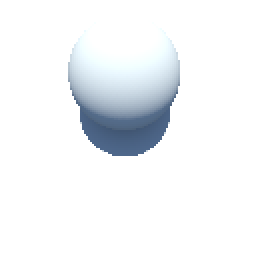}  & 
\includegraphics[height=.065\linewidth]{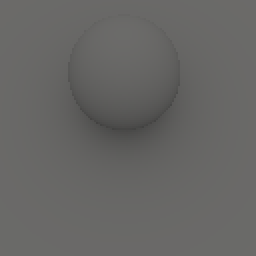}  &
\includegraphics[height=.065\linewidth]{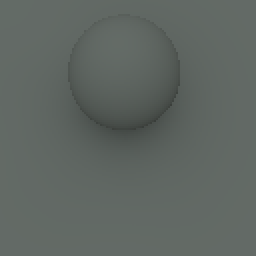}  & 
\includegraphics[height=.065\linewidth]{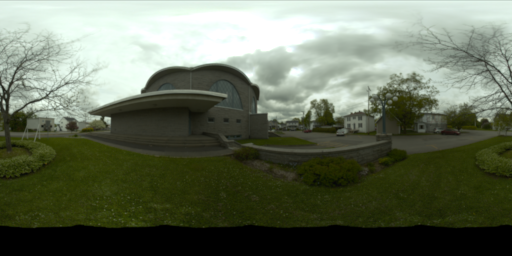}    & 
\includegraphics[height=.065\linewidth]{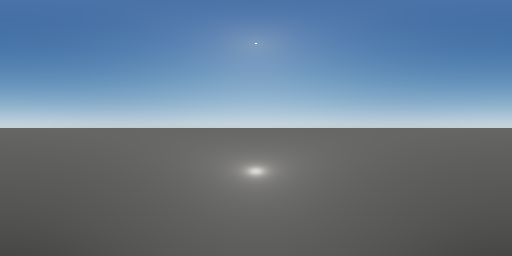}  & 
\includegraphics[height=.065\linewidth]{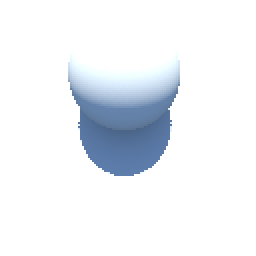}  & 
\includegraphics[height=.065\linewidth]{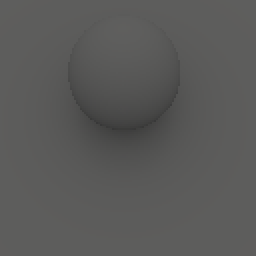}  &
\includegraphics[height=.065\linewidth]{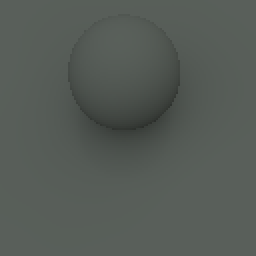}  \\ 
\rotatebox{90}{30th} & 
\includegraphics[height=.065\linewidth]{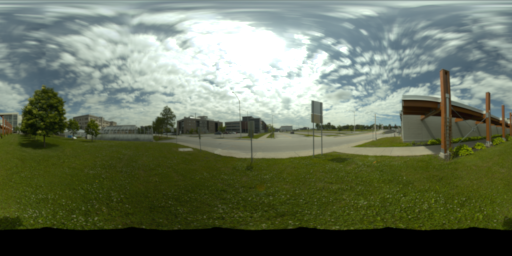}    & 
\includegraphics[height=.065\linewidth]{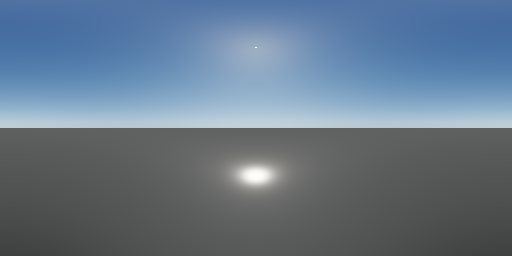}  & 
\includegraphics[height=.065\linewidth]{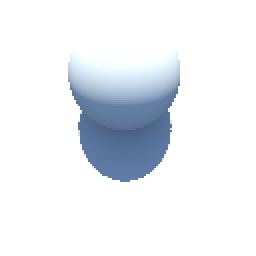}  & 
\includegraphics[height=.065\linewidth]{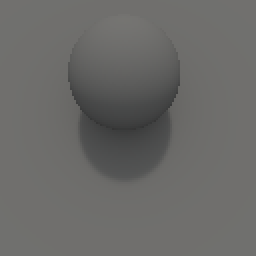}  &
\includegraphics[height=.065\linewidth]{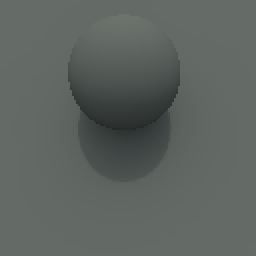}  & 
\includegraphics[height=.065\linewidth]{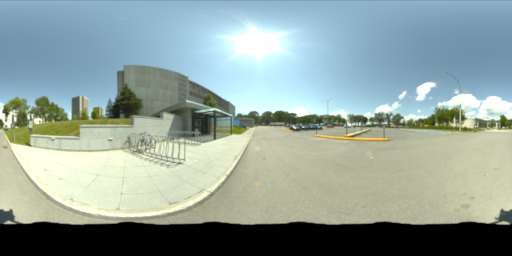}    & 
\includegraphics[height=.065\linewidth]{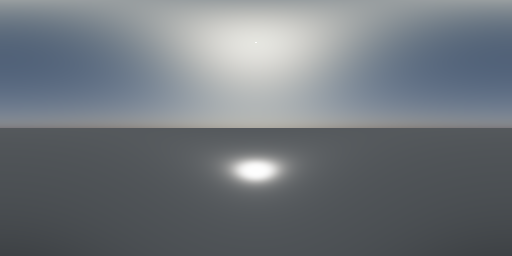}  & 
\includegraphics[height=.065\linewidth]{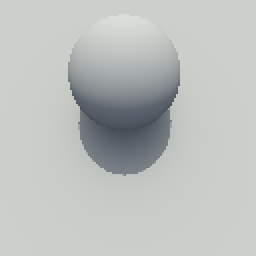}  & 
\includegraphics[height=.065\linewidth]{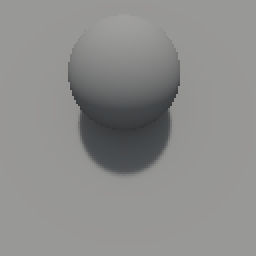}  &
\includegraphics[height=.065\linewidth]{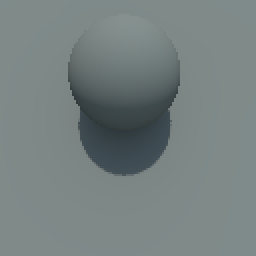}  \\ 
\rotatebox{90}{50th} & 
\includegraphics[height=.065\linewidth]{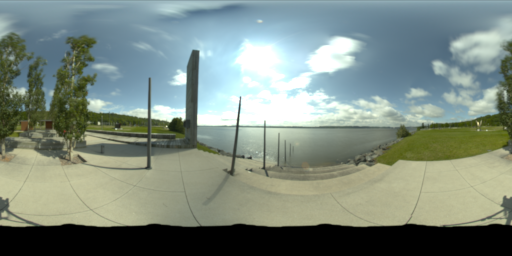}    & 
\includegraphics[height=.065\linewidth]{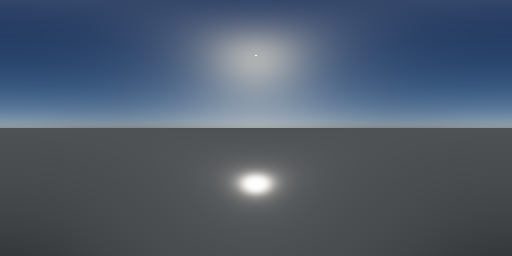}  & 
\includegraphics[height=.065\linewidth]{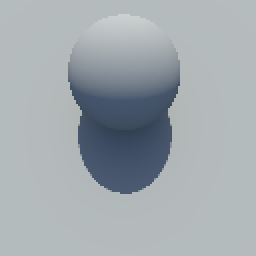}  & 
\includegraphics[height=.065\linewidth]{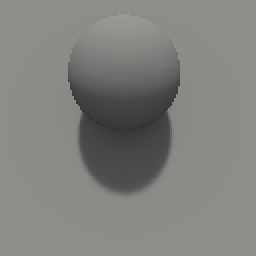}  &
\includegraphics[height=.065\linewidth]{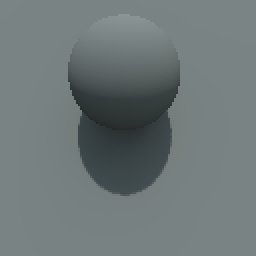}  & 
\includegraphics[height=.065\linewidth]{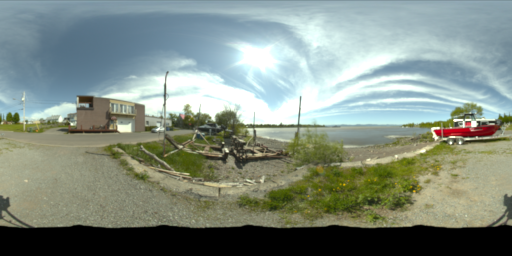}    &  
\includegraphics[height=.065\linewidth]{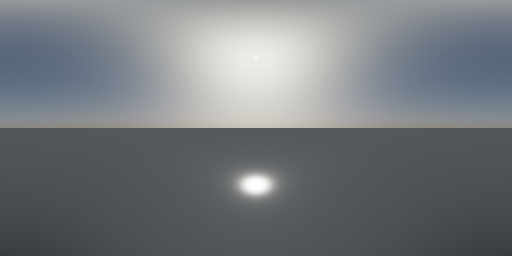}  & 
\includegraphics[height=.065\linewidth]{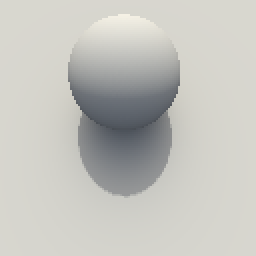}  & 
\includegraphics[height=.065\linewidth]{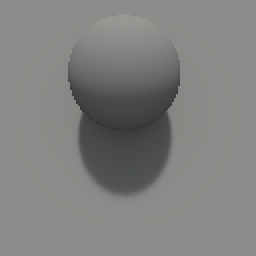}  &
\includegraphics[height=.065\linewidth]{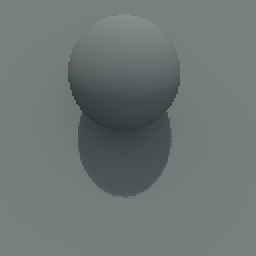}  \\ 
\rotatebox{90}{80th} & 
\includegraphics[height=.065\linewidth]{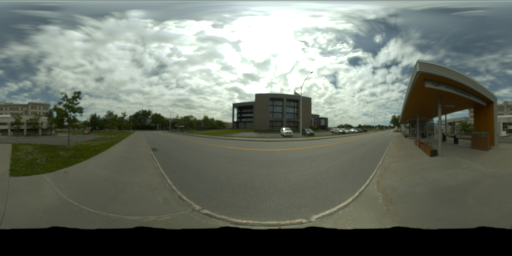}    & 
\includegraphics[height=.065\linewidth]{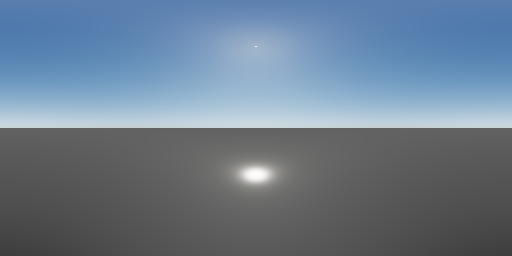}  & 
\includegraphics[height=.065\linewidth]{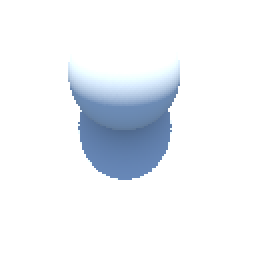}  & 
\includegraphics[height=.065\linewidth]{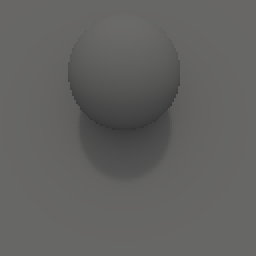}  &
\includegraphics[height=.065\linewidth]{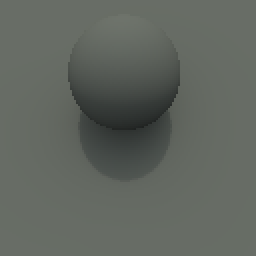}  & 
\includegraphics[height=.065\linewidth]{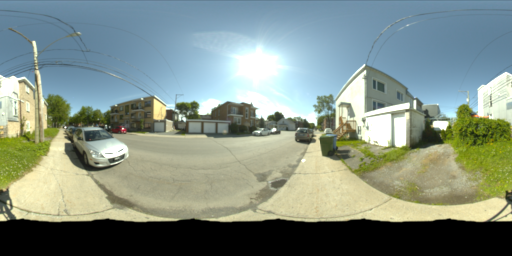}    & 
\includegraphics[height=.065\linewidth]{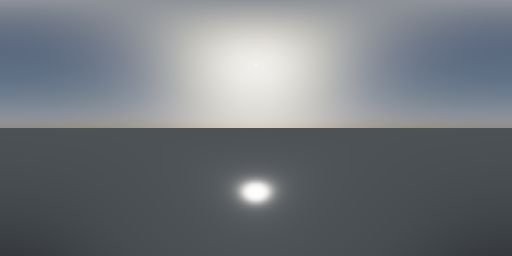}  & 
\includegraphics[height=.065\linewidth]{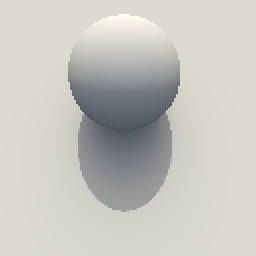}  & 
\includegraphics[height=.065\linewidth]{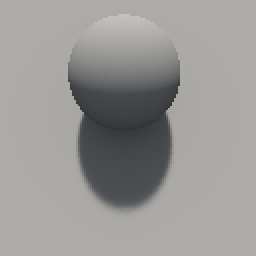}  &
\includegraphics[height=.065\linewidth]{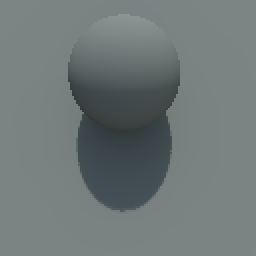}  \\ 
& 
(a) LDR panorama & \makecell{(b) HW sky (top) \\ LM sky (bottom)} & (c) HW~\cite{holdgeoffroy-cvpr-17} & \makecell{(d) Ours \\ (PanoNet)} & (e) GT & 
(a) LDR panorama & \makecell{(b) HW sky (top) \\ LM sky (bottom)} & (c) HW~\cite{holdgeoffroy-cvpr-17} & \makecell{(d) Ours \\ (PanoNet)} & (e) GT \\
\end{tabular}
\vspace{.5em}
\caption{Qualitative comparison between approaches to estimate HDR parametric lighting from a single LDR panorama. We compare our PanoNet method, which learns to predict the parameters of the LM model~\cite{lalonde-3dv-14}, to the approach of \cite{holdgeoffroy-cvpr-17} which {directly fits} the parameters of the HW model~\cite{hosek-siggraph-12,hosek-cga-13} using non-linear optimization. For each example, we show (a) the input LDR panorama, (b) the reconstructed skies using both the HW (top) and LM (bottom) models, (c--e) a rendering of a simple scene (viewed from the top) using \cite{holdgeoffroy-cvpr-17}, ours, and the ground truth respectively. The rows are ordered by si-RMSE percentile for our technique (see fig.~\ref{fig:pano.boxplot} for the overall error distribution). {All the renders are tone-mapped with a $\gamma=2.2$ for visualization purposes}. Our PanoNet can better fit different weather conditions such that the renders are similar to the real lighting both in terms of intensity and shadow softness. Please zoom in for details.} 
\label{fig:pano.fitting}
\end{figure*}

\subsection{Evaluating the PanoNet CNN}

To evaluate the ability of the PanoNet network to predict the LM sky parameters from a single LDR panorama, we employ 103 outdoor HDR panoramas~\cite{yannick-cvpr-19}. We first convert each panorama to LDR by applying a random exposure factor to the HDR panorama, clipping its maximum value at 1, and quantizing to 8 bits. To quantify performance, the scale-invariant (si-)RMSE error metric is computed by rendering the diffuse sphere+plane scene (sec.~\ref{sec:panonet-loss-functions}) with the ground truth panorama (original HDR), the results of the non-linear fitting approach of \cite{holdgeoffroy-cvpr-17}, and our PanoNet. For the scene, we use a simple sphere placed on a ground plane, viewed from the top. We place the sun of both methods to its ground truth position---this ensures that sun position is factored out in the evaluation. 

\begin{figure}[t]
\centering
\setlength{\tabcolsep}{0pt}

\begin{tabular}{cccc}
\includegraphics[height=.4\linewidth, trim={0 0 .5cm 0},clip]{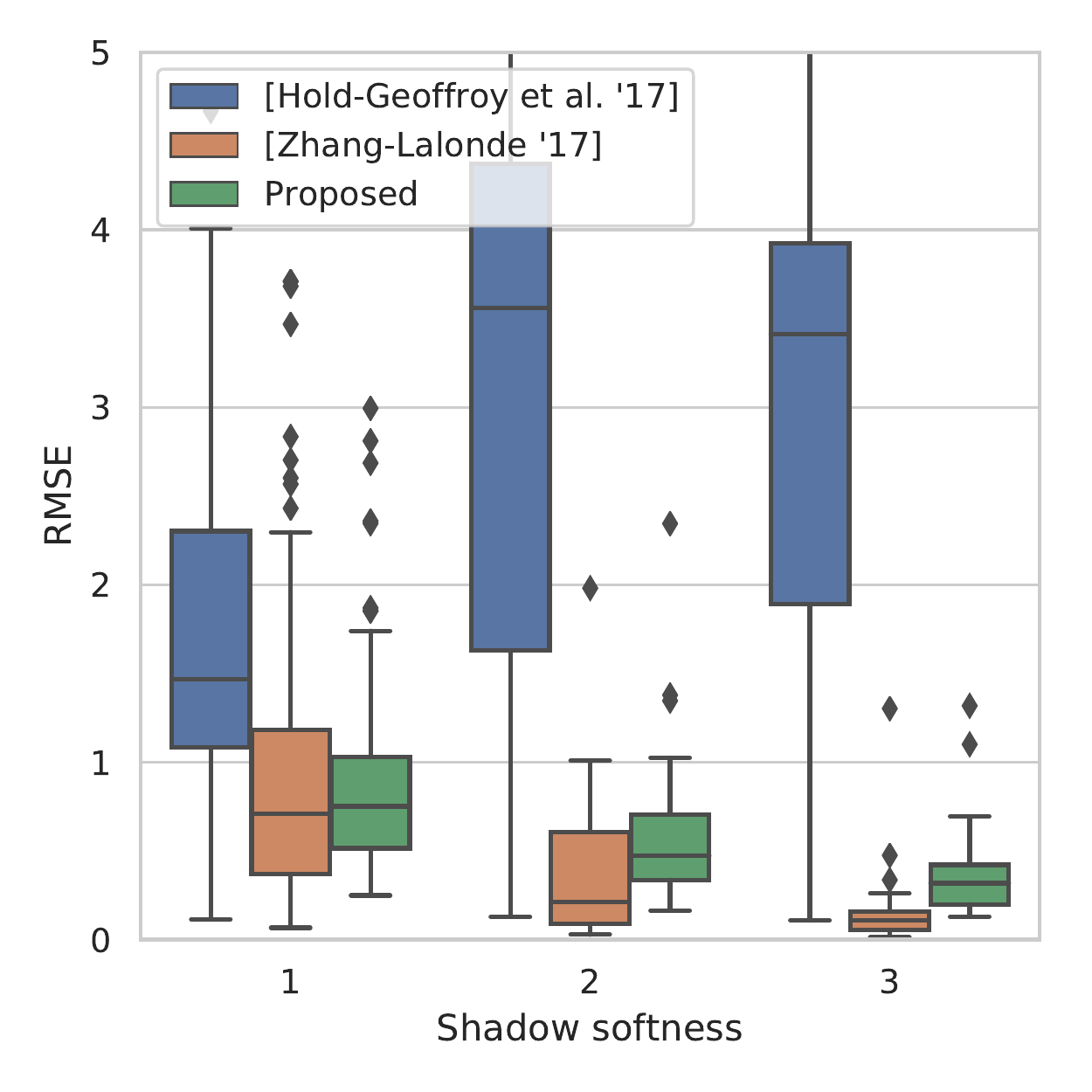} &
\includegraphics[height=.4\linewidth, trim={0.95cm 0.45cm 0.6cm 0},clip]{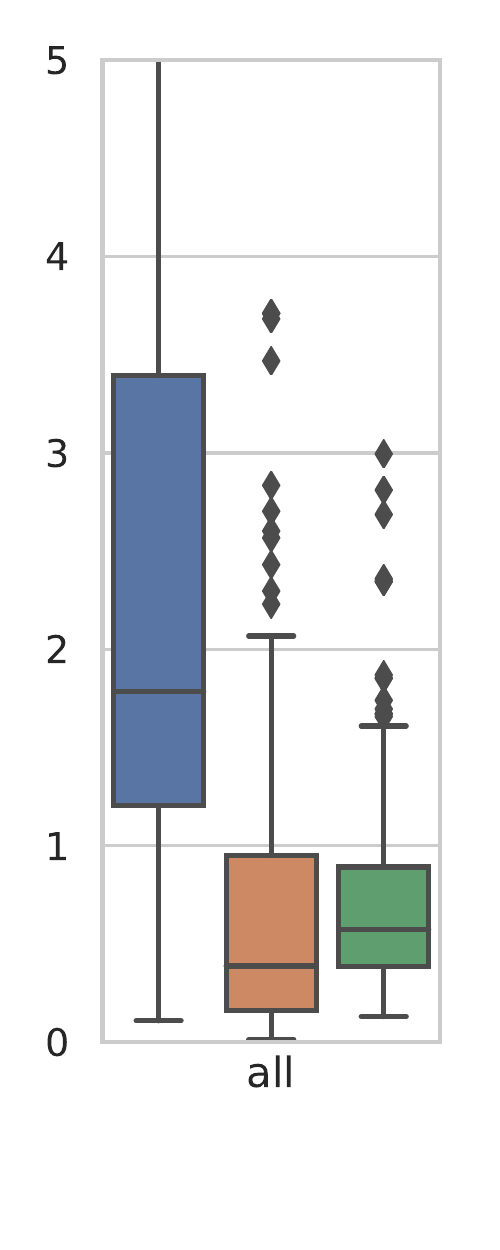} &
\includegraphics[height=.4\linewidth, trim={0 0 .5cm 0},clip]{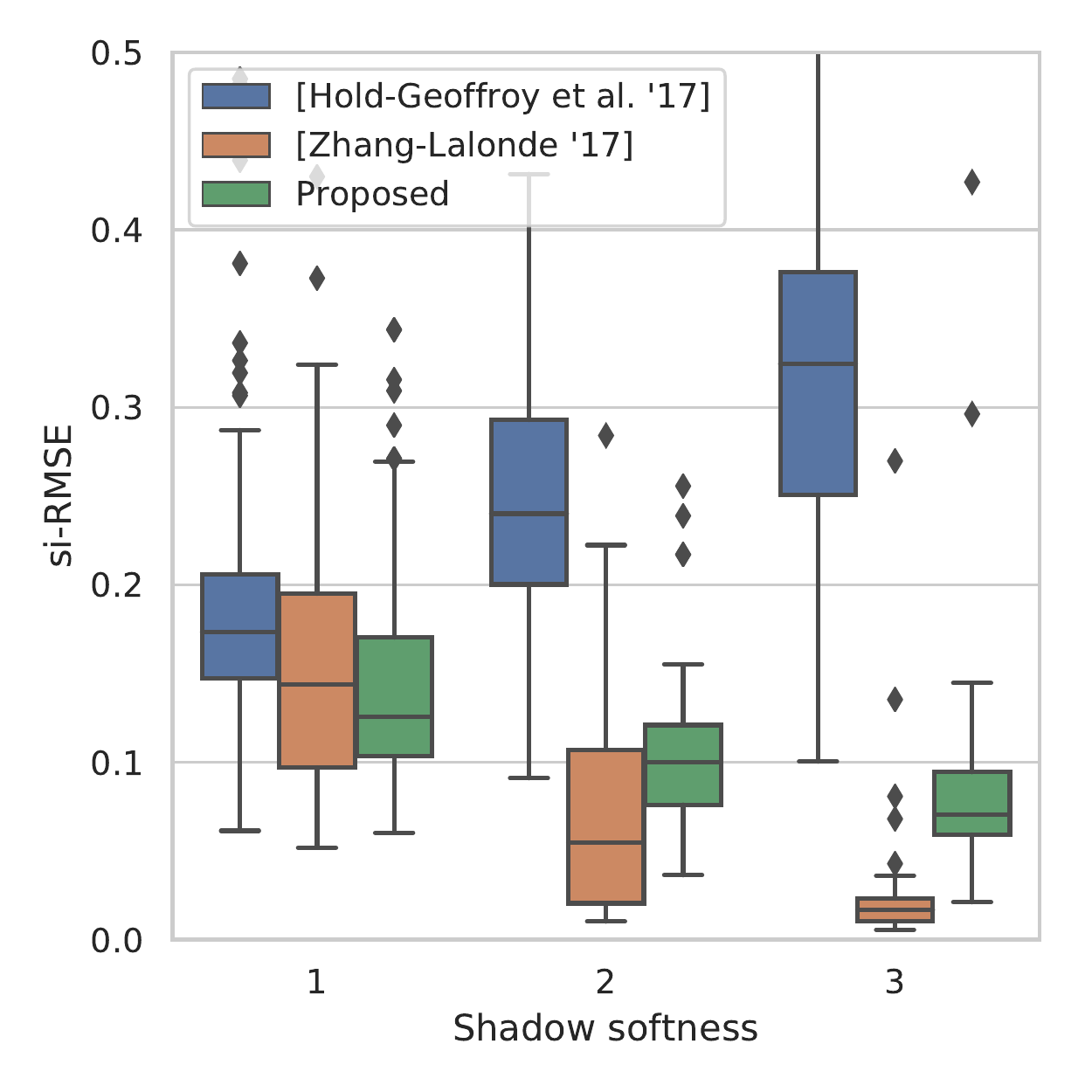} &
\includegraphics[height=.4\linewidth, trim={1.2cm .22cm 0 0},clip]{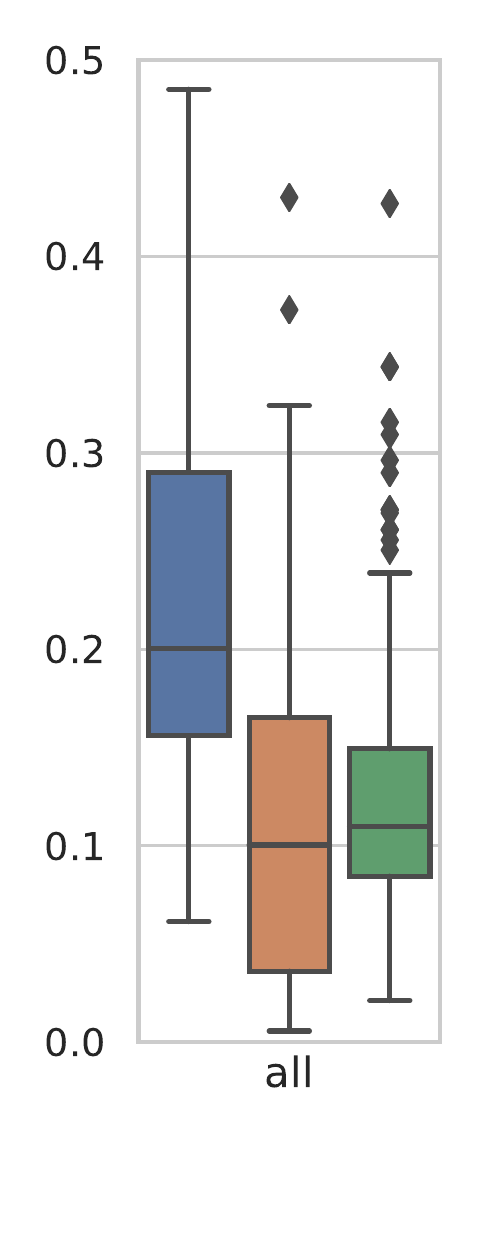} \\ 
\end{tabular}
\caption{Quantitative comparison of panorama fitting on the real dataset. We show a distribution of (left) RMSE and (right) scale-invariant RMSE~\cite{grosse-iccv-09} as a function of the shadow softness, ranging from very sharp (1) to smooth (3) (see text for the shadow softness definition used), and over the entire dataset (``all''). Compared to the previous state-of-the-art~\cite{holdgeoffroy-cvpr-17}, our PanoNet provides better fits to the wide range of illumination conditions. {The non-parametric approach of \cite{zhang-iccv-2017} provides a lower bound on the prediction.} The lower (upper) edge of each box indicates the 25th (75th) percentiles.
}
\label{fig:pano.boxplot}
\end{figure}

{We show some qualitative visual examples in fig.~\ref{fig:pano.fitting}; these are accompanied by the quantitative comparison results shown in fig.~\ref{fig:pano.boxplot}}. To better highlight the different lighting conditions in the HDR panorama dataset, we split it into 3 categories based on the softness of shadows in the scene. Shadow softness is estimated by first computing the histogram of horizontal gradients on a 5-pixel row immediately below the sphere. Then, a reference sunny image is manually identified, and the KL-divergence between the histograms of each image and that of the reference is computed. Finally, the dataset is sorted by the KL-divergence, and subsequently split into three groups by empirically determining cutting points where shadows appear sharp (corresponding to a softness of 1, or sunny skies) and where shadows are not visible (corresponding to a softness of 3, or overcast skies). The remainder of the dataset is classified as having a softness of 2 (partially sunny skies). As shown in fig.~\ref{fig:pano.boxplot}, {the non-parametric model \cite{zhang-iccv-2017} is unsurprisingly more precise than both parametric models, however it does not produce intuitive sky parameters that can be used to subsequently train CropNet.} The approach of \cite{holdgeoffroy-cvpr-17} yields low errors when shadows are sharp (sunny skies), but the error significantly increases when shadows should be softer (partially cloudy to overcast skies). In contrast, our PanoNet performs similarly well across all lighting conditions.

{
We compare two variants of PanoNet: trained with our proposed combined loss (eqs~\ref{eqn:ldr2hdr} and \ref{eqn:lm}) and with the parametric loss only (eq.~\ref{eqn:lm}). The performance at estimating lighting in different weather conditions is shown in table~\ref{tab:Ablation}, and indicates that adding the additional losses from~\cite{zhang-iccv-2017} helps PanoNet to encode more information in the latent space. 
}
\begin{table}[t]
\vspace{0em}
\centering
\small
\begin{tabular}{llcccc}
\toprule
\multicolumn{2}{c}{\makecell{Shadow softness \\ weather}}  & \makecell{1 \\ clear} & \makecell{2 \\ mixed} & \makecell{3 \\ cloudy} & all \\
\midrule
\multirow{2}{*}{RMSE} & eq.~\ref{eqn:lm} only & 1.07 & 0.55 & 0.59 & 0.82 \\
                      & eqs~\ref{eqn:ldr2hdr} and \ref{eqn:lm}	& \textbf{0.92}  & \textbf{0.41} & \textbf{0.36} & \textbf{0.73} \\
\midrule
\multirow{2}{*}{si-RMSE} & eq.~\ref{eqn:lm} only & 0.19 & 0.13 & \textbf{0.08} & 0.16\\
                        & eqs~\ref{eqn:ldr2hdr} and \ref{eqn:lm}	& \textbf{0.15}  & \textbf{0.11} & 0.09 & \textbf{0.13} \\
\bottomrule
\end{tabular}
\vspace{.5em}
\caption{Ablation study comparing the use of different loss functions (eq.~\ref{eqn:lm} only vs. using both eqs~\ref{eqn:ldr2hdr} and \ref{eqn:lm}) when training PanoNet. }
\label{tab:Ablation}
\vspace{-1em}
\end{table}

\subsection{Evaluating the CropNet CNN}

\begin{figure}[!h]
\footnotesize
\centering
\includegraphics[width=0.8\linewidth, trim={0 0 0 0},clip]{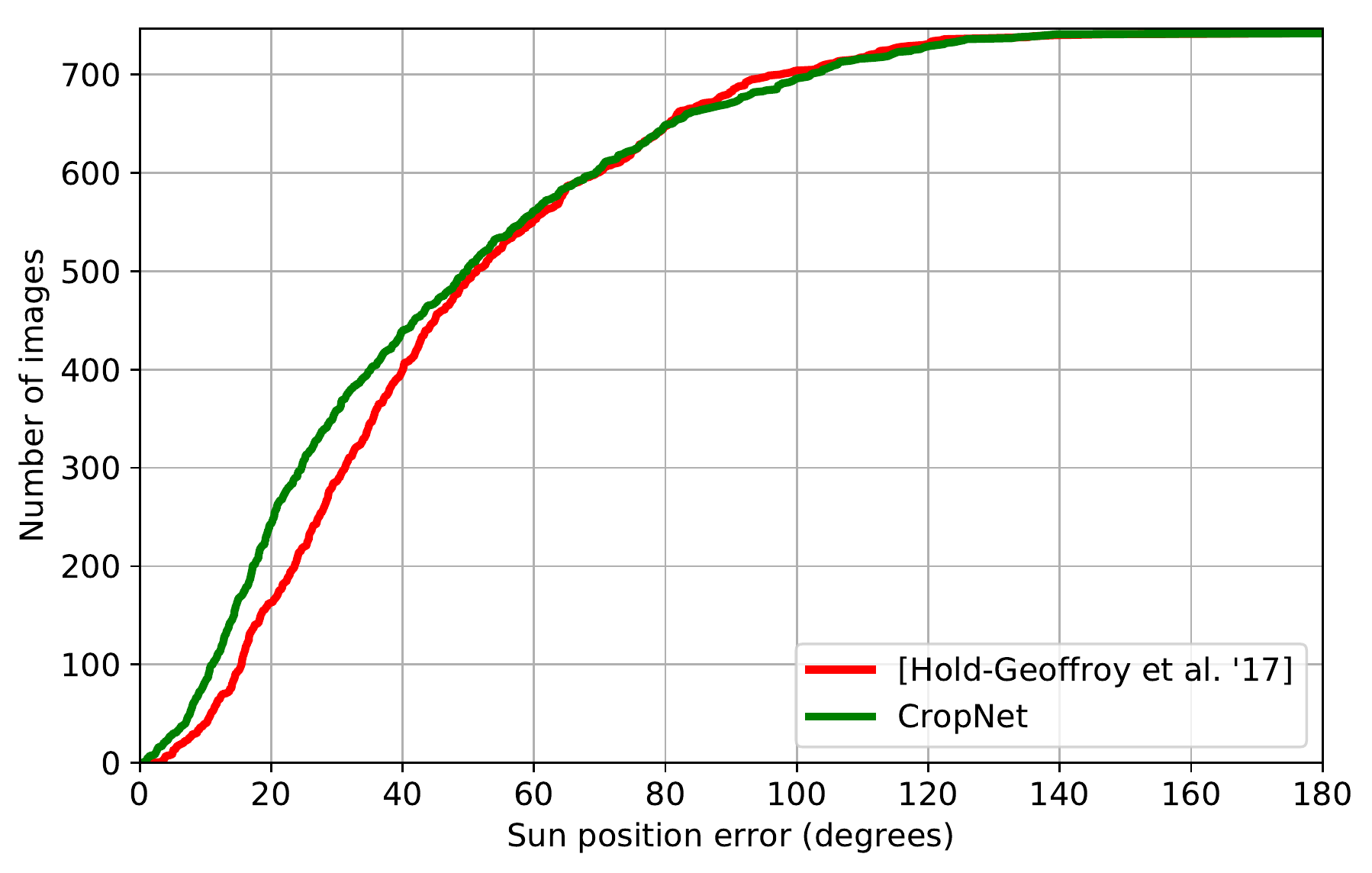}
\vspace{-1em}
\caption{Cumulative sun angular error comparison between our CropNet and~\cite{holdgeoffroy-cvpr-17} on single images extracted from SUN360 dataset~\cite{xiao-cvpr-12}. Our method slightly outperforms that of~\cite{holdgeoffroy-cvpr-17}.
}
\label{fig:crop.sunpos}
\end{figure}

We extract 7 limited field-of-view photos from each panorama in the HDR outdoor panorama dataset~\cite{yannick-cvpr-19} using a standard pinhole camera model, and randomly sampling the camera azimuth. Then, we estimate lighting parameters from these crops using the approach of \cite{holdgeoffroy-cvpr-17} and our CropNet. 
{The sun position angular error from our CropNet and~\cite{holdgeoffroy-cvpr-17} is shown in fig.~\ref{fig:crop.sunpos}. Note that the sun position is independent from the other radiometric lighting parameters in our approach. In contrast, the radiometric sky properties are constrained by the sun position in~\cite{holdgeoffroy-cvpr-17}. Therefore, in order to fairly compare the radiometric lighting parameters, we employ the network of~\cite{holdgeoffroy-cvpr-17} to estimate the sun position in the subsequent experiments.}

\begin{figure}[!t]
\footnotesize
\centering 
\setlength{\tabcolsep}{1pt}
\begin{tabular}{ccccc}
\includegraphics[height=.22\linewidth]{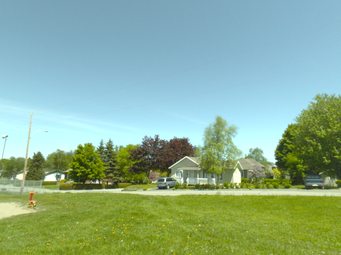}  & 
\includegraphics[height=.22\linewidth]{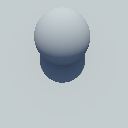}  & 
\includegraphics[height=.22\linewidth]{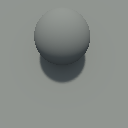} &
\includegraphics[height=.22\linewidth]{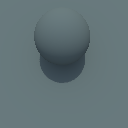}  \\
\includegraphics[height=.22\linewidth]{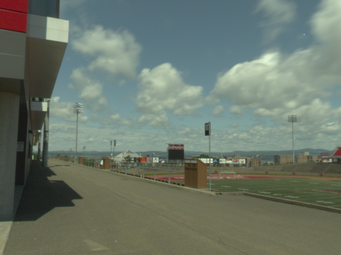}  & 
\includegraphics[height=.22\linewidth]{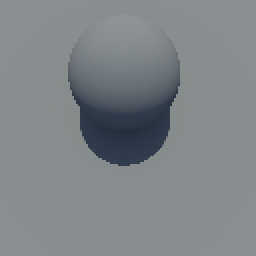}  & 
\includegraphics[height=.22\linewidth]{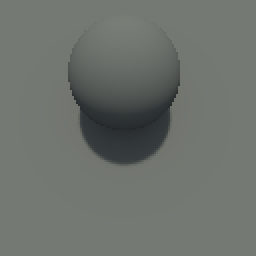} &
\includegraphics[height=.22\linewidth]{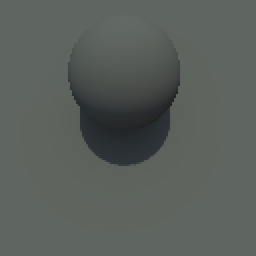}  \\
\includegraphics[height=.22\linewidth]{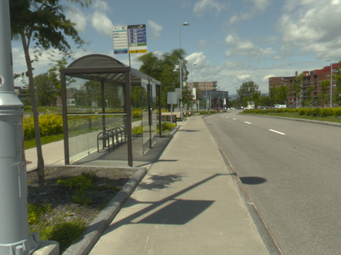}  & 
\includegraphics[height=.22\linewidth]{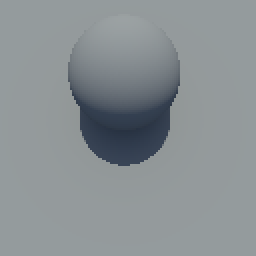}  & 
\includegraphics[height=.22\linewidth]{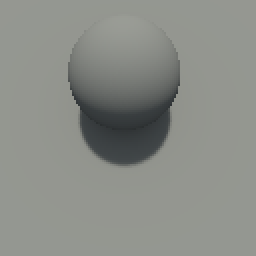} &
\includegraphics[height=.22\linewidth]{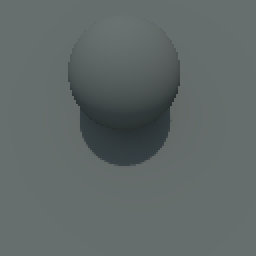}  \\
\includegraphics[height=.22\linewidth]{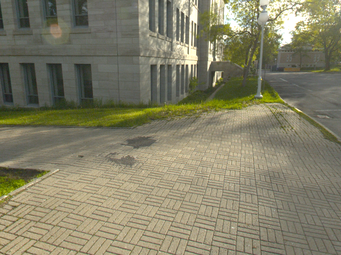}  & 
\includegraphics[height=.22\linewidth]{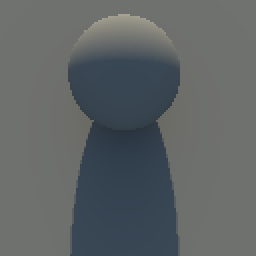}  & 
\includegraphics[height=.22\linewidth]{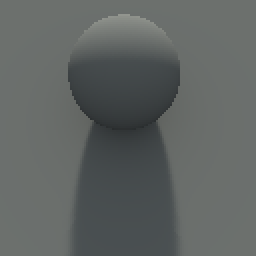} &
\includegraphics[height=.22\linewidth]{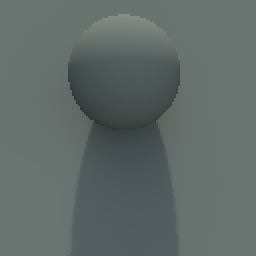}  \\
\includegraphics[height=.22\linewidth]{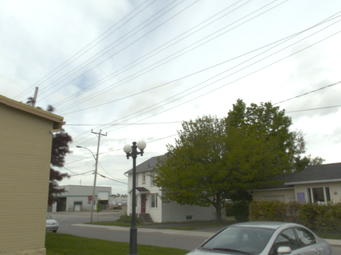}  & 
\includegraphics[height=.22\linewidth]{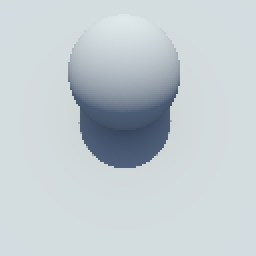}  & 
\includegraphics[height=.22\linewidth]{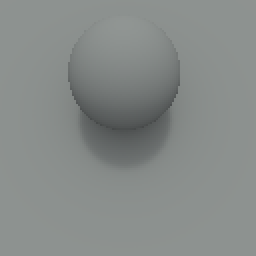} &
\includegraphics[height=.22\linewidth]{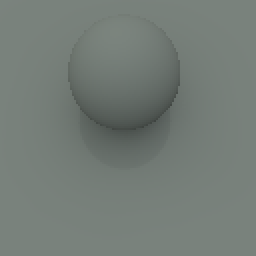}  \\ 
\includegraphics[height=.22\linewidth]{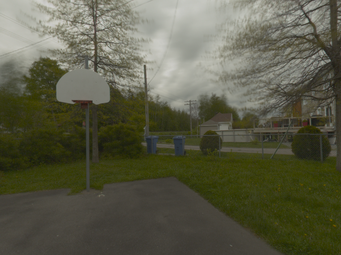}  & 
\includegraphics[height=.22\linewidth]{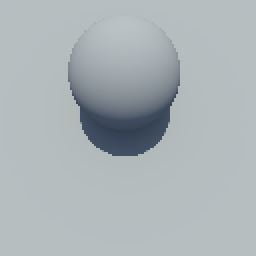}  & 
\includegraphics[height=.22\linewidth]{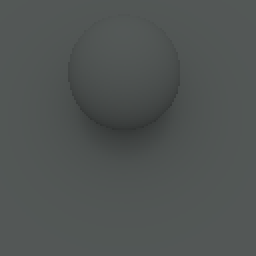} &
\includegraphics[height=.22\linewidth]{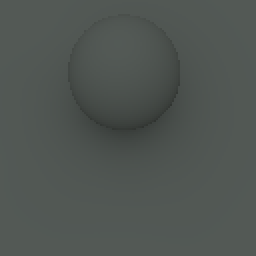}  \\
(a) Input & (b) ~\cite{holdgeoffroy-cvpr-17} &(c) CropNet & (d) GT render
\end{tabular}
\vspace{.5em}
\caption[]{Lighting estimation comparison with ground truth. We crop an image from an HDR panorama database~\cite{yannick-cvpr-19}, then the cropped image (a) is used to estimate the lighting parameters from \cite{holdgeoffroy-cvpr-17} and CropNet. The renders from each approach are shown in (b) and (c); (d) shows the ground truth render from the HDR panorama.
\vspace{-1em}} 
\label{fig:crop.lavaloutdoor}
\end{figure}

Fig.~\ref{fig:crop.lavaloutdoor} shows qualitative comparison between the prediction from \cite{holdgeoffroy-cvpr-17} and our CropNet with the ground truth. Again, our method can accurately estimate lighting conditions ranging from clear to overcast. For example, in an overcast day (last row of fig.~\ref{fig:crop.lavaloutdoor}), our approach can successfully estimate the lighting and produce renders with soft shadows. However, we notice that the approach in~\cite{holdgeoffroy-cvpr-17} constantly outputs a clear sky, and usually fails to generate soft shadows.

\begin{figure}[!h]
\footnotesize
\centering
\setlength{\tabcolsep}{0pt}
\begin{tabular}{cccc}
\includegraphics[height=.4\linewidth, trim={0 0 .5cm 0},clip]{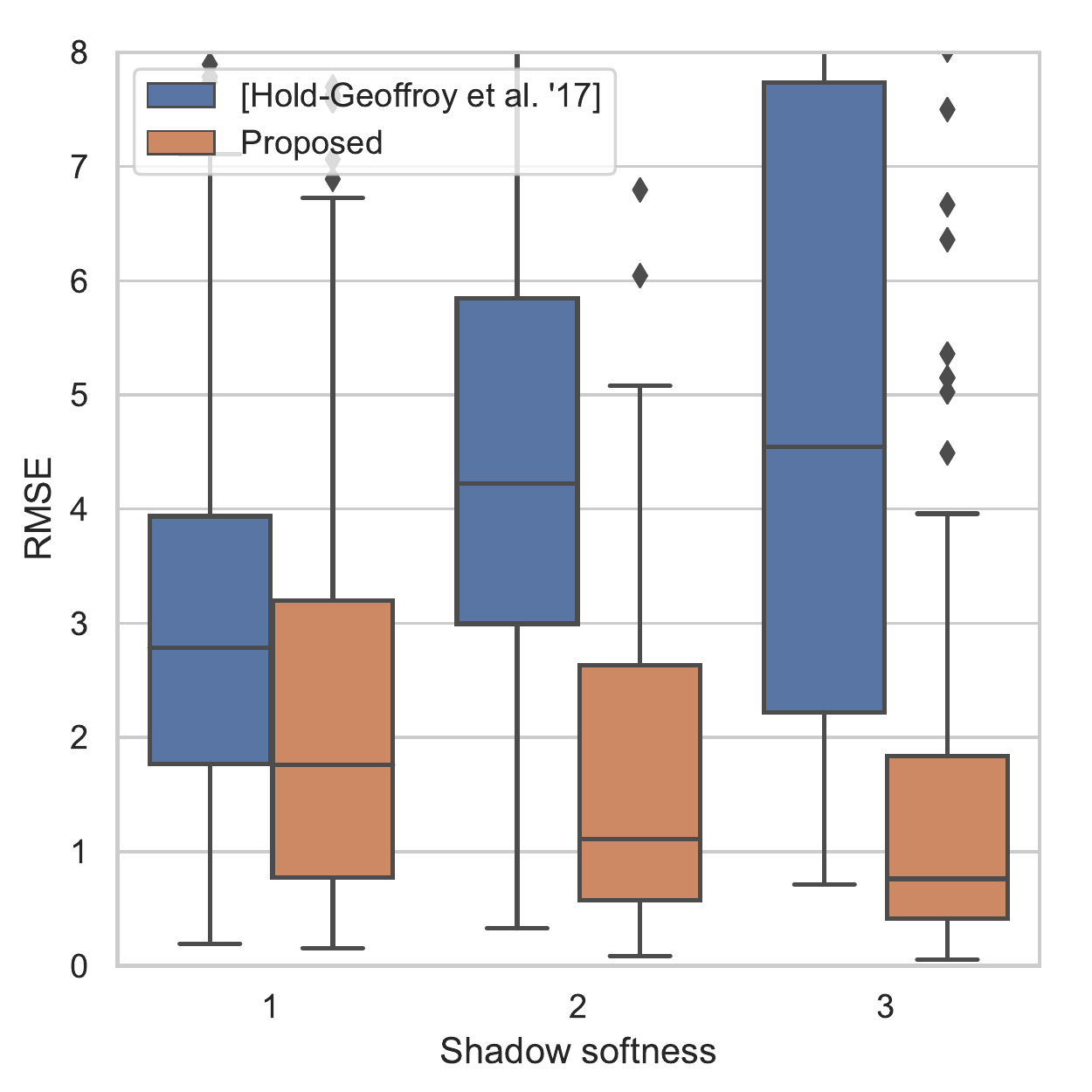} &
\includegraphics[height=.4\linewidth, trim={0.95cm 0.45cm 0.6cm 0},clip]{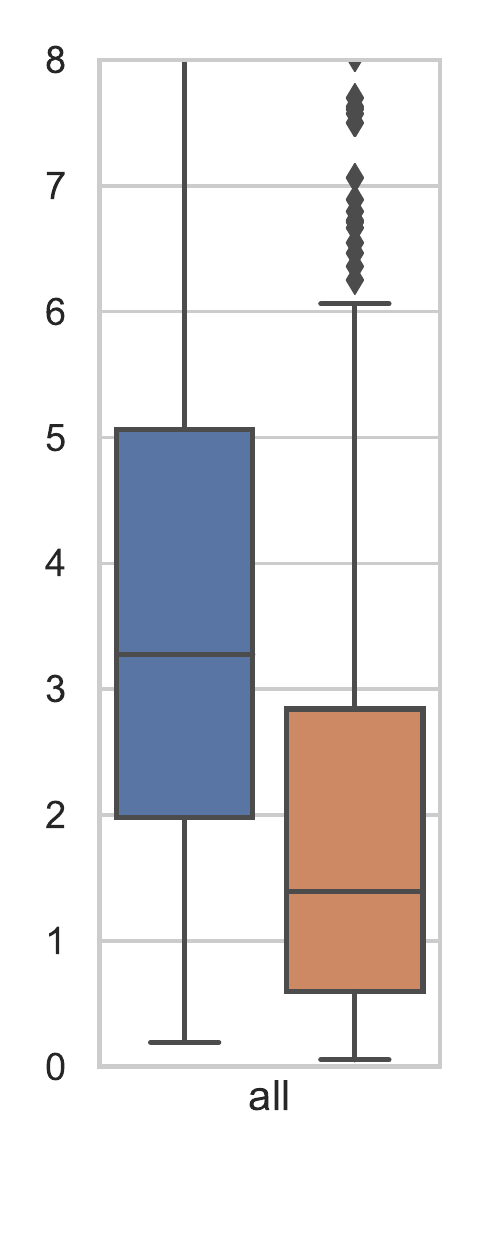} &
\includegraphics[height=.4\linewidth, trim={0 0 .5cm 0},clip]{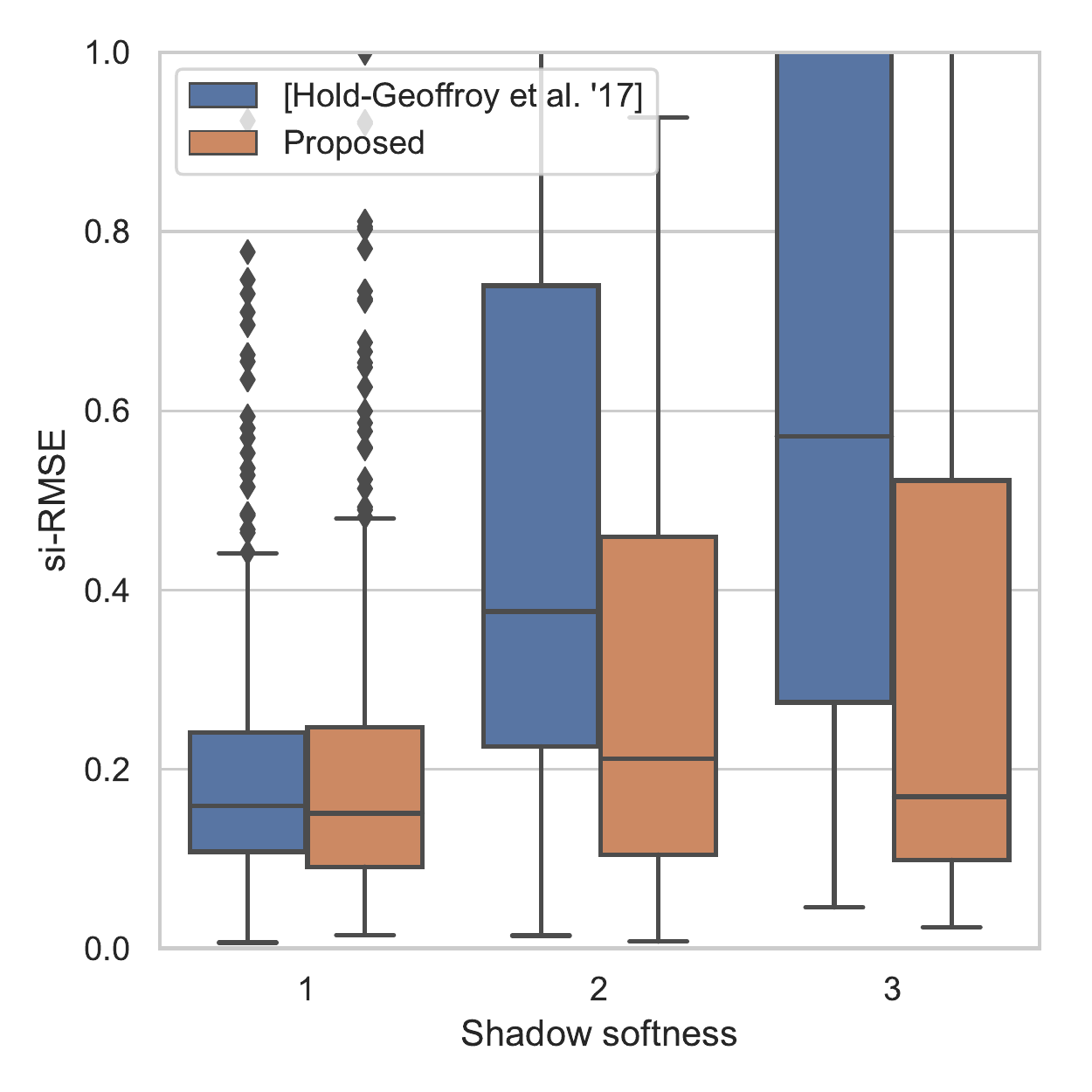} &
\includegraphics[height=.4\linewidth, trim={1.2cm .22cm 0 0},clip]{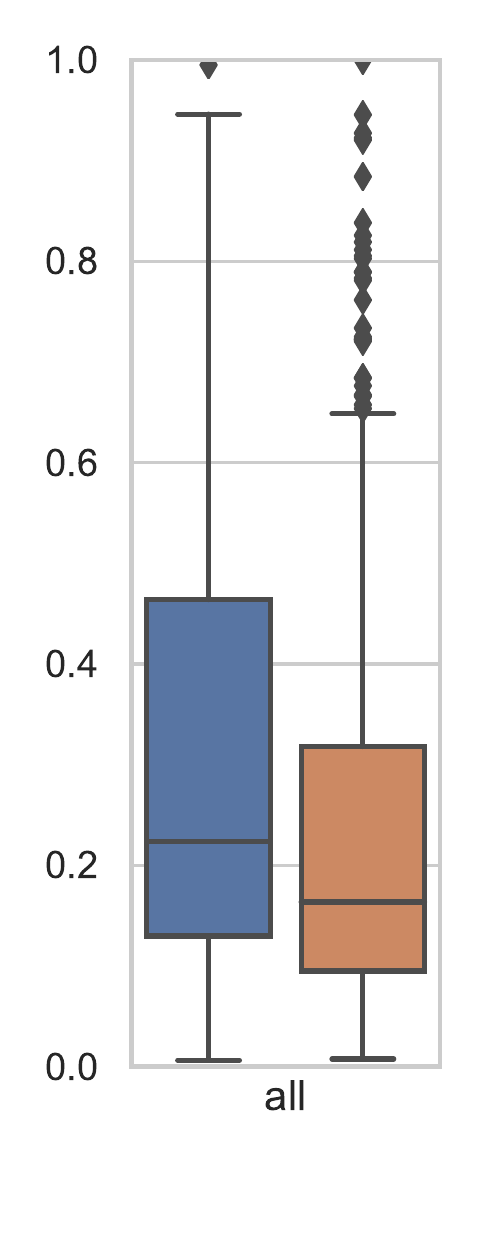} \\ 
\end{tabular}
\vspace{.25em}
\caption{Quantitative comparison between our CropNet and \cite{holdgeoffroy-cvpr-17} on single images extracted from a dataset of HDR panoramas~\cite{yannick-cvpr-19}, where the ground truth lighting is available. The render error is shown in the box-plots as function of shadow softness ranging from very sharp (1) to smooth (3) (see text for the exact definition), and over the entire dataset (``all''). The lower (upper) edge of each box indicates the 25th (75th) percentiles. While both techniques perform relatively similarly when shadows are very sharp (in sunny conditions), the error of \cite{holdgeoffroy-cvpr-17} increases when the sky is not completely clear and shadows start to disappear. In contrast, our method remains much more stable. 
\vspace{-1em}
}
\label{fig:crop.boxplot}
\end{figure}

Those qualitative results are validated quantitatively in fig.~\ref{fig:crop.boxplot}, which reports both the RMSE and si-RMSE metrics with respect to renders obtained with the ground truth lighting. Again, our approach shows much improved performance across different weather conditions and error metrics. The RMSE plot demonstrates that our CropNet can obtain a much more accurate estimate of exposure in outdoor scenes. The si-RMSE shows that the estimated lighting generates more faithful shadows when compared to \cite{holdgeoffroy-cvpr-17}.

\begin{figure}[!t]
\footnotesize
\centering 
\setlength{\tabcolsep}{2pt}
\begin{tabular}{ccc}
\includegraphics[height=.28\linewidth]{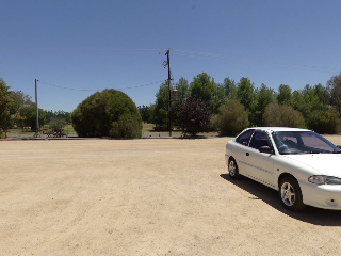}  & 
\includegraphics[height=.28\linewidth]{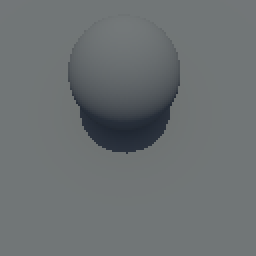}   & 
\includegraphics[height=.28\linewidth]{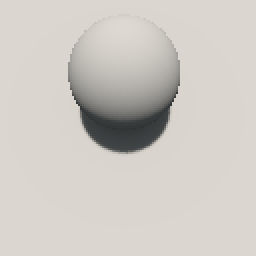} \\

\includegraphics[height=.28\linewidth]{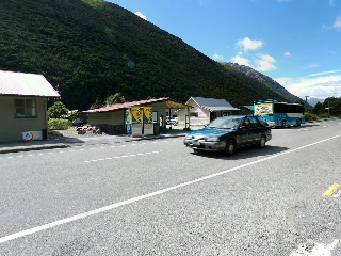}  & 
\includegraphics[height=.28\linewidth]{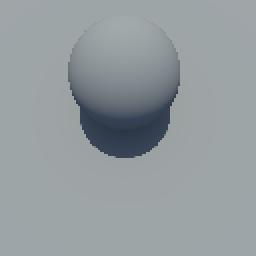}   & 
\includegraphics[height=.28\linewidth]{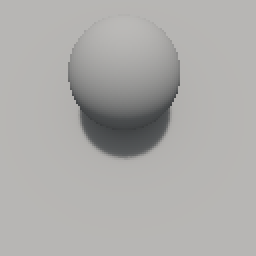} \\

\includegraphics[height=.28\linewidth]{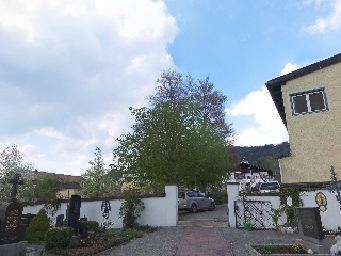}  & 
\includegraphics[height=.28\linewidth]{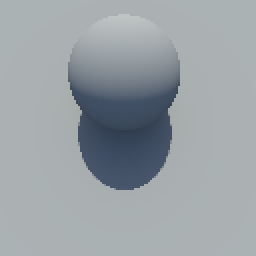}  & 
\includegraphics[height=.28\linewidth]{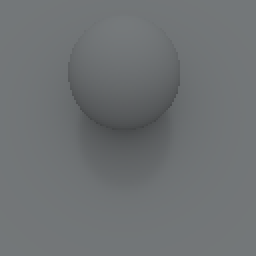} \\

\includegraphics[height=.28\linewidth]{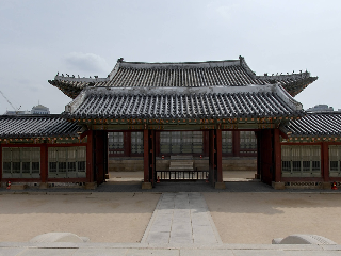}  & 
\includegraphics[height=.28\linewidth]{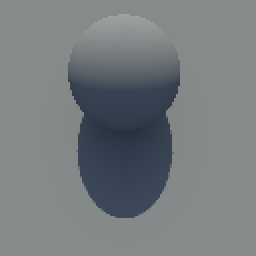}  & 
\includegraphics[height=.28\linewidth]{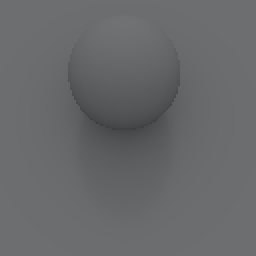} \\

\includegraphics[height=.28\linewidth]{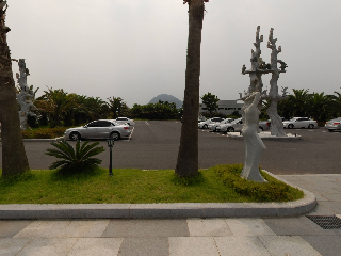}  & 
\includegraphics[height=.28\linewidth]{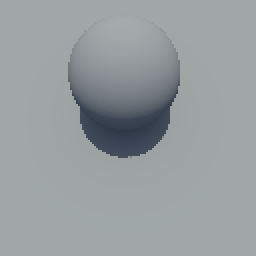}  & 
\includegraphics[height=.28\linewidth]{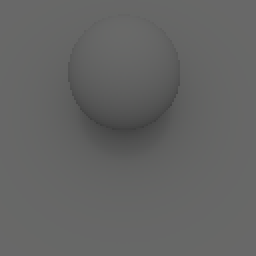} \\

\includegraphics[height=.28\linewidth]{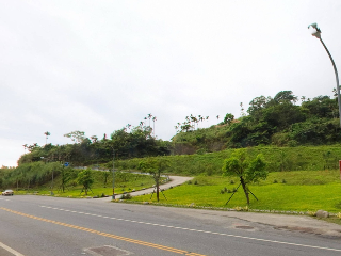}  & 
\includegraphics[height=.28\linewidth]{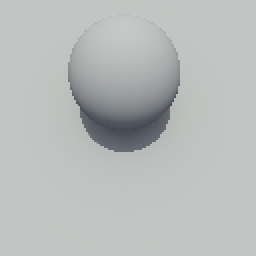}  & 
\includegraphics[height=.28\linewidth]{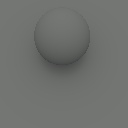} \\

(a) Input image &
(b) \cite{holdgeoffroy-cvpr-17} & 
(c) Ours (CropNet)
\end{tabular}
\caption{Lighting estimation for real images. We show different lighting estimation results from real images. Our approach provides consistent estimations in a wide range of illumination conditions, ranging from clear (top) to overcast (bottom). Here, the sun azimuth is kept fixed to better compare the renders.
\vspace{-1em}}  
\label{fig:crop.sun360}
\end{figure}

\begin{figure}[!t]
\footnotesize
\centering 
\setlength{\tabcolsep}{2pt}
\begin{tabular}{lccc}
\rotatebox{90}{\hspace{.5cm}Original panorama} & 
\multicolumn{3}{c}{
\includegraphics[width=.785\linewidth]{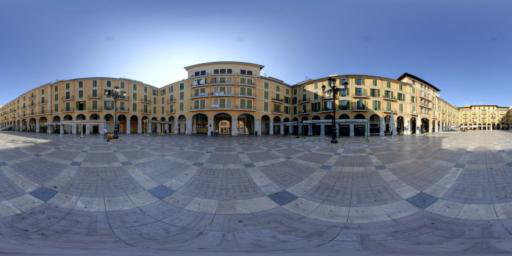}}  \\
\rotatebox{90}{\hspace{.4cm}Crops} & 
\includegraphics[width=.25\linewidth]{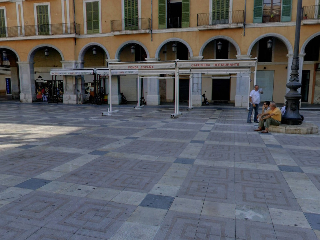}  &
\includegraphics[width=.25\linewidth]{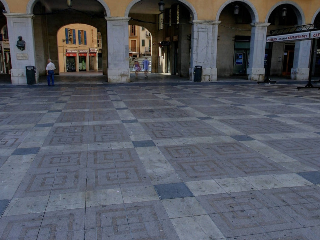}  & 
\includegraphics[width=.25\linewidth]{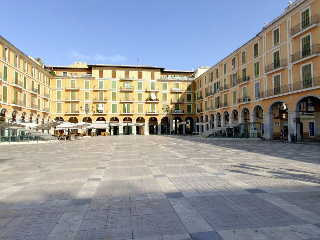}  \\ 
\rotatebox{90}{\hspace{.15cm}Ours (CropNet)} & 
\includegraphics[width=.25\linewidth]{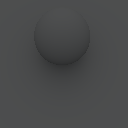}  &
\includegraphics[width=.25\linewidth]{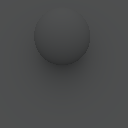}  &  
\includegraphics[width=.25\linewidth]{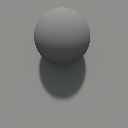}  \\ 
\rotatebox{90}{\hspace{.9cm}\cite{holdgeoffroy-cvpr-17}} & 
\includegraphics[width=.25\linewidth]{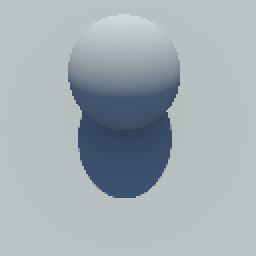}  &
\includegraphics[width=.25\linewidth]{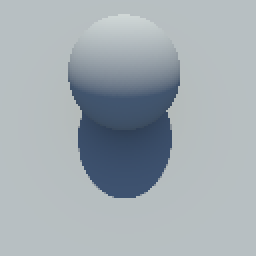}  & 
\includegraphics[width=.25\linewidth]{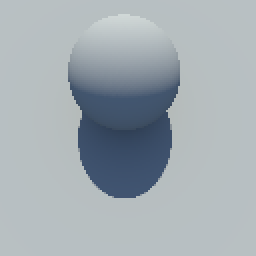}  \\ 
\end{tabular}
\caption{Estimating lighting for different viewpoints extracted from the same panorama. From a single panorama (top row), we extract three different crops from left to right (2nd row), and compare the lighting estimates obtained with our method (3rd row) to that of \cite{holdgeoffroy-cvpr-17} (bottom row). Despite being extracted from the same panorama, the crops exhibit different lighting conditions. Our method adapts to these changes naturally, and predicts believable illumination conditions in all three cases. Here, the sun azimuth is kept fixed to better compare the renders.
\vspace{-1em}} 
\label{fig:crop.viewpoints}
\end{figure}

{Fig.~\ref{fig:crop.sun360} shows qualitative comparison between our CropNet with~\cite{holdgeoffroy-cvpr-17} for real images. Our approach is able to handle different weather conditions ranging from fully sunny to overcast skies. Fig.~\ref{fig:crop.viewpoints} compares the lighting between our CropNet and~\cite{holdgeoffroy-cvpr-17} for different viewpoints extracted from the same panorama. Our approach better adapts the \emph{local} crop lighting than~\cite{holdgeoffroy-cvpr-17}. More examples can be found in the supplementary material.}

\section{Discussion}

In this paper, we presented a method for estimating HDR lighting from a single, LDR image. At the heart of our approach lies a CNN that learns to predict the parameters of an analytical sky model from a single LDR panorama such that it 1) more realistically reconstructs the appearance of the sky, and 2) renders the appearance of objects lit by this illumination. This CNN is used to label a large dataset of outdoor panoramas, which is in turn used to train a second CNN, this time to estimate the lighting parameters from a single, limited field of view image. Due to its intuitive set of lighting parameters such as the sun shape and color, our approach is particularly amenable to applications where a user might want to modify the estimated lighting parameters, either because they were judged to be not quite right, or to experiment with different lighting effects. In addition, the network outputs can be used to render an environment map, which can readily be used to insert photorealistic objects into photographs (fig.~\ref{fig:teaser}). 

While our approach outperforms the state-of-the-art both qualitatively and quantitatively, it is not without limitations. Its most noticeable limitation is that it typically has difficulty in properly identifying soft shadows, both from panoramas and from crops. We suspect this is because the HDR training data does not contain many examples where this is the case. In addition, since shadows are dimmer, they are harder to see in the images, as such cues to their existence are most subtle. Another limitation is the tendency to estimate gray skies, even when the sky in the image is visible, and blue. Again, we suspect that this is a data issue: the network has difficulty in overcoming the fact that most panoramas have clouds which bring the average sky color closer to gray. 
These questions pave the way for exciting future work.


\section*{Acknowledgements}
This work was partially supported by the REPARTI Strategic Network and the NSERC Discovery Grant RGPIN-2014-05314. We gratefully acknowledge the support of Nvidia with the donation of the GPUs used for this research, as well as Adobe for generous gift funding.

{\small
\bibliographystyle{ieee}
\bibliography{refs}
}

\end{document}